\documentclass[sigconf]{acmart}
\AtBeginDocument{%
  }

\copyrightyear{2026}
\acmYear{2026}
\setcopyright{cc}
\setcctype{by}
\acmConference[MM '26]{Proceedings of the 34th ACM International Conference on Multimedia}{November 10--14, 2026}{Rio de Janeiro, Brazil}
\acmBooktitle{Proceedings of the 34th ACM International Conference on Multimedia (MM '26), November 10--14, 2026, Rio de Janeiro, Brazil}
\acmDOI{10.1145/3767308.3835712}
\acmISBN{979-8-4007-2213-4/2026/11}

\usepackage{graphicx}
\usepackage{booktabs}
\usepackage{enumitem}
\usepackage{makecell}
\usepackage{pifont}
\usepackage{multirow}
\newcommand{\cmark}{\ding{51}} % ✓
\newcommand{\xmark}{\ding{55}} % ✗
\usepackage{enumitem}
\usepackage{balance}
\usepackage[ruled,linesnumbered,vlined]{algorithm2e}

\usepackage{amsmath,amssymb}
\usepackage{listings}

\begin{document}

%%
%% The "title" command has an optional parameter,
%% allowing the author to define a "short title" to be used in page headers.
\title{BooM-VVT: Boosting Mask-Free Video Virtual Try-On with Image-Level Pseudo Data}

\settopmatter{
  % printacmref=true,
  authorsperrow=4
}

%%
%% The "author" command and its associated commands are used to define
%% the authors and their affiliations.
%% Of note is the shared affiliation of the first two authors, and the
%% "authornote" and "authornotemark" commands
%% used to denote shared contribution to the research.
\author{Wei Zhang}
\orcid{0009-0001-6054-8719}
\affiliation{%
  \institution{Nanjing University of Science and Technology}
  \city{Nanjing}
  \state{Jiangsu}
  \country{China}
}
\email{zwplus\_pro@njust.edu.cn}

% 3. Co-Author
\author{Xin Li}
\orcid{0000-0002-6352-6523}
\affiliation{%
  \institution{University of Science and Technology of China}
  \city{Hefei}
  \state{Anhui}
  \country{China}
}
\email{xin.li@ustc.edu.cn}

% 2. Co-Author
\author{Peishu Shi}
\affiliation{%
  \institution{National University of Singapore}
  \city{Singapore}
  \country{Singapore}
}
\email{peishushi@u.nus.edu}

% 4. Co-Author
\author{Jialin Gao}
\orcid{0000-0002-8554-7827}
\affiliation{%
  \institution{Meituan}
  \city{Beijing}
  \country{China}
}
\email{gao.linge@gmail.com}

% 5. Co-Author
\author{Xuekang Peng}
\orcid{0009-0006-7499-2227}
\affiliation{%
  \institution{Nanjing University of Science and Technology}
  \city{Nanjing}
  \state{Jiangsu}
  \country{China}
}
\email{pengxuekang@njust.edu.cn}

% 6. Co-Author
\author{Zhichao Lian}
\correspondingauthor
\authornote{Zhichao Lian and Yeying Jin are co-corresponding authors.}
\affiliation{%
  \institution{Nanjing University of Science and Technology}
  \city{Nanjing}
  \state{Jiangsu}
  \country{China}
}
\email{lzcts@163.com}

% 7. Co-Author
\author{Yeying Jin}
\correspondingauthor
\authornotemark[1]
\authornote{Project Leader.}
\orcid{0000-0001-7818-9534}
\affiliation{%
  \institution{National University of Singapore}
  \city{Singapore}
  \country{Singapore}
}
\email{jinyeying@u.nus.edu}

% Running header for a paper with six or more authors
\renewcommand{\shortauthors}{Wei Zhang et al.}

%%
%% The abstract is a short summary of the work to be presented in the
%% article.
\begin{abstract}
Video virtual try-on (VVT) aims to generate realistic videos of a person wearing a target garment. Recent methods leverage a keyframe-driven video generation paradigm to improve in-the-wild performance, yet they still rely on masks to localize try-on regions, making them vulnerable to large motions and severe occlusions. Although mask-free image-based try-on methods have shown promising results by leveraging large-scale pseudo data, extending this paradigm to videos remains difficult, as constructing video-level pseudo data is prohibitively expensive. Furthermore, coarse keyframe sampling and the scarcity of multi-view try-on data limit existing keyframe-driven methods in maintaining garment consistency and handling diverse try-on tasks.

To address these challenges, we propose BooM-VVT, a mask-free VVT framework built upon the keyframe-driven paradigm. To achieve mask-free VVT, we introduce a multi-stage training strategy that leverages image-level pseudo data for mask-free localization learning, substantially reducing the need for costly video-level pseudo data. To improve garment consistency, we propose Garment-Sensitive Keyframe Sampling, which selects keyframes based on garment-relevant body regions to better capture garment appearance. We further introduce Frame-Shared 3D-RoPE to establish spatiotemporal correspondences between keyframes and target video frames for accurate garment-detail transfer. Finally, we construct OmniView, a large-scale multi-view try-on dataset to support reliable try-on video generation under complex camera viewpoints and diverse try-on tasks. Extensive experiments demonstrate that BooM-VVT achieves superior temporal consistency and garment fidelity over existing methods. Project page: \url{https://boomvvt.github.io/boomvvt}.
\end{abstract}

%%
%% The code below is generated by the tool at http://dl.acm.org/ccs.cfm.
%% Please copy and paste the code instead of the example below.
%%


%%
%% Keywords. The author(s) should pick words that accurately describe
%% the work being presented. Separate the keywords with commas.
%% A "teaser" image appears between the author and affiliation
%% information and the body of the document, and typically spans the
%% page.
\begin{teaserfigure}
  \includegraphics[width=\textwidth]{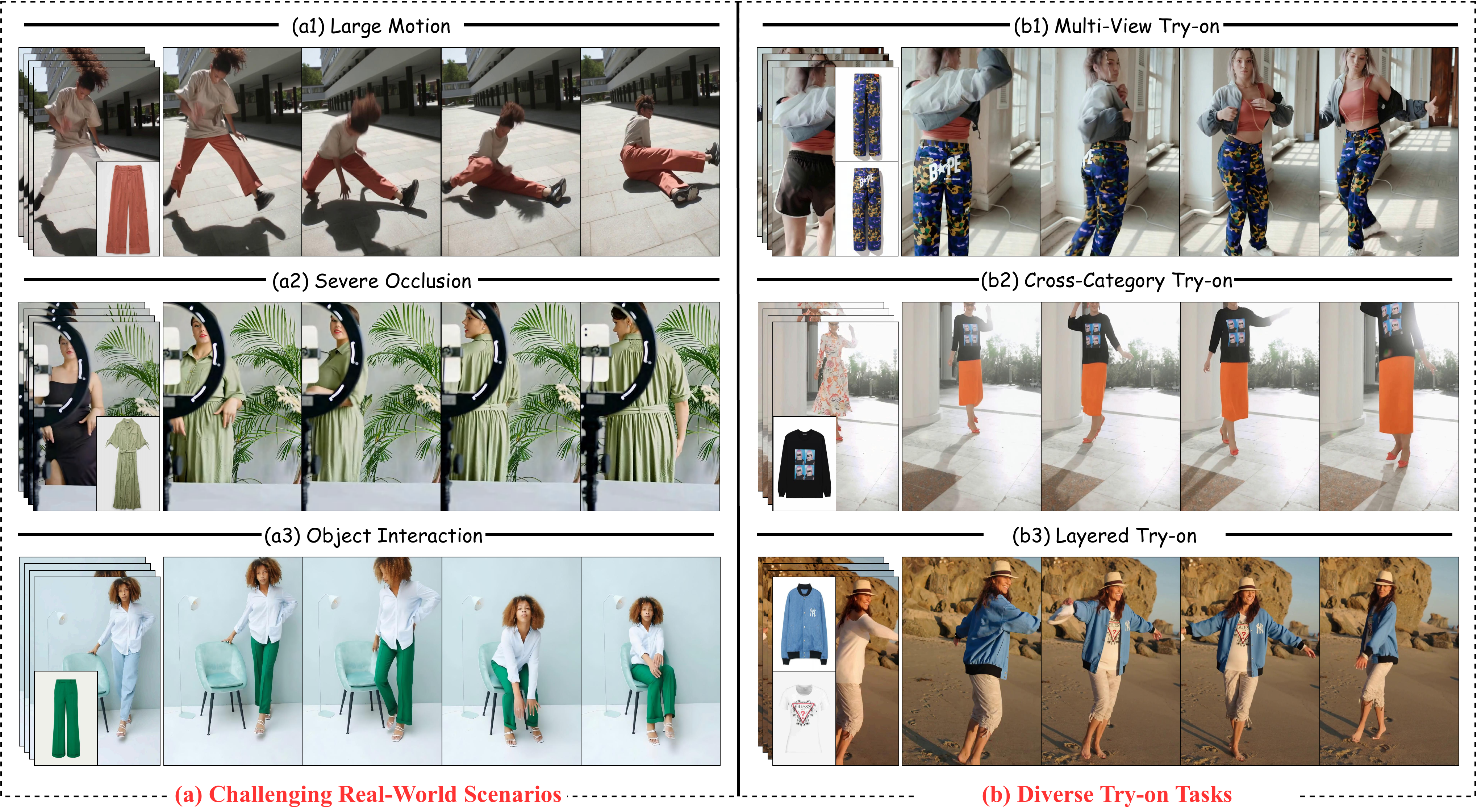}
  \caption{BooM-VVT generates high-fidelity and temporally coherent virtual try-on videos without requiring masks. The left panel shows its robustness in challenging real-world scenarios, including large motion, severe occlusion, and object interaction. The right panel highlights its flexibility in diverse try-on tasks, such as multi-view, cross-category, and layered try-on.
}
\Description{The teaser contains two groups of virtual try-on examples. The left group shows source video frames, target garment references, and generated try-on frames in scenes involving large body motion, severe occlusion, and object interaction. The right group shows multi-view, cross-category, and layered try-on examples, with person inputs, garment references, and the corresponding generated results.}
  \label{fig:teaser}
\end{teaserfigure}

%%
%% This command processes the author and affiliation and title
%% information and builds the first part of the formatted document.
\maketitle

\section{Introduction}
Video virtual try-on (VVT) aims to synthesize realistic videos of a person wearing a target garment, with broad applications in e-commerce and digital content creation. Although recent diffusion-based methods~\cite{li2025magictryon,chong2025catv2ton,zeng2025eevee,nguyen2025swifttry,pan2025once,karras2024fashion} have achieved impressive progress, they remain heavily dependent on scarce paired garment--video data. Due to the limited scale and scene diversity of existing datasets~\cite{fang2024vivid,dong2019fw}, these methods often struggle to preserve garment fidelity and temporal consistency in real-world scenarios.

To address this limitation, recent studies~\cite{he2025devil,zuo2025dreamvvt} have introduced a keyframe-driven try-on video generation paradigm that decomposes the end-to-end VVT pipeline into two stages: (1) generating try-on images for selected keyframes, and (2) generating the final try-on video conditioned on the resulting keyframe try-on images. Under this paradigm, methods such as DreamVVT~\cite{zuo2025dreamvvt} can exploit large-scale unpaired video data to train the video generation model, thereby improving generalization in real-world scenarios. Despite these advantages, keyframe-driven methods still face several  challenges. 
\textbf{First}, existing VVT methods typically rely on explicit masks to localize try-on regions. However, accurately estimating masks in videos is highly challenging due to large pose variations and severe occlusions, as shown in Fig.~\ref{fig:teaser}(a). Inaccurate masks often introduce temporal artifacts such as jittering and flickering. While mask-free image-based try-on methods~\cite{zhang2025boow,zhang2026unifit,guo2025any2anytryon,feng2025omnitry} have shown promising results by leveraging large-scale pseudo data, extending this paradigm to video virtual try-on remains nontrivial, as constructing video-level pseudo data is prohibitively expensive. 
\textbf{Second}, existing methods typically select keyframes based on overall pose changes and full-body visibility. However, this strategy is often influenced by body regions irrelevant to the target garment, such as the lower body when trying on tops. As a result, the selected keyframes may fail to fully capture the garment appearance, thereby compromising garment consistency in the synthesized videos.
\textbf{Third}, existing multi-view try-on datasets remain limited in scale and coverage, as shown in Table~\ref{tab:dataset_compare}. This limitation hinders current methods from generating plausible results under challenging viewpoints (e.g., back views) and handling diverse try-on tasks (see Fig.~\ref{fig:teaser}(b)).

To tackle these challenges, we present BooM-VVT, a novel mask-free video virtual try-on framework built upon the keyframe-driven paradigm. To achieve mask-free VVT, we propose a multi-stage training strategy that enables the model to leverage more accessible image-level pseudo data to learn robust mask-free try-on region localization. As a result, the model can adapt to mask-free video virtual try-on with only minimal video-level pseudo data, substantially easing data construction. To improve garment consistency, we first propose a Garment-Sensitive Keyframe Sampling (GSKS) strategy. Specifically, it selects keyframes based on body regions most relevant to the target garment, allowing the selected keyframes to better capture the target garment appearance.
We further introduce Frame-Shared 3D-RoPE. By assigning identical positional encodings to keyframe try-on images and their corresponding target video frames, it explicitly establishes spatiotemporal correspondences between them, enabling more accurate garment appearance transfer. Finally, to address the scarcity of multi-view try-on data, we construct OmniView, a large-scale multi-view try-on dataset that enables reliable try-on generation across diverse viewpoints and try-on tasks. In summary, our contributions are as follows: 
\begin{itemize}
\item We propose BooM-VVT, a mask-free video virtual try-on framework that leverages a multi-stage training strategy to learn robust try-on region localization from readily available image-level pseudo data, substantially reducing reliance on costly video-level pseudo data.
\item To improve garment consistency, we propose a Garment-Sensitive Keyframe Sampling (GSKS) strategy and Frame-Shared 3D-RoPE. GSKS selects more informative keyframes based on garment-relevant body regions, while Frame-Shared 3D-RoPE facilitates more accurate garment appearance transfer by establishing spatiotemporal correspondences between the keyframes and the target video.
\item We construct OmniView, a large-scale multi-view try-on dataset that enables reliable try-on generation across diverse viewpoints and try-on tasks.
\end{itemize}

\begin{table}[t]
\caption{Comparison with existing multi-view try-on datasets. \textit{Category Diversity} indicates whether the dataset covers diverse garment categories, and \textit{Garment Completeness} indicates whether complete flat garment images are provided.}
\label{tab:dataset_compare}
\small
\centering
\begin{tabular}{l c c c}
\toprule
Dataset & \makecell{Garment Category\\Diversity} & \makecell{Garment\\Completeness} & Scale \\
\midrule
MVG & \xmark & \xmark & 1,009 \\
\textbf{OmniView} & \cmark & \cmark & 6,110 \\
\bottomrule
\end{tabular}
\end{table}

\begin{figure*}[t]
  \centering
  \includegraphics[width=0.9\textwidth]{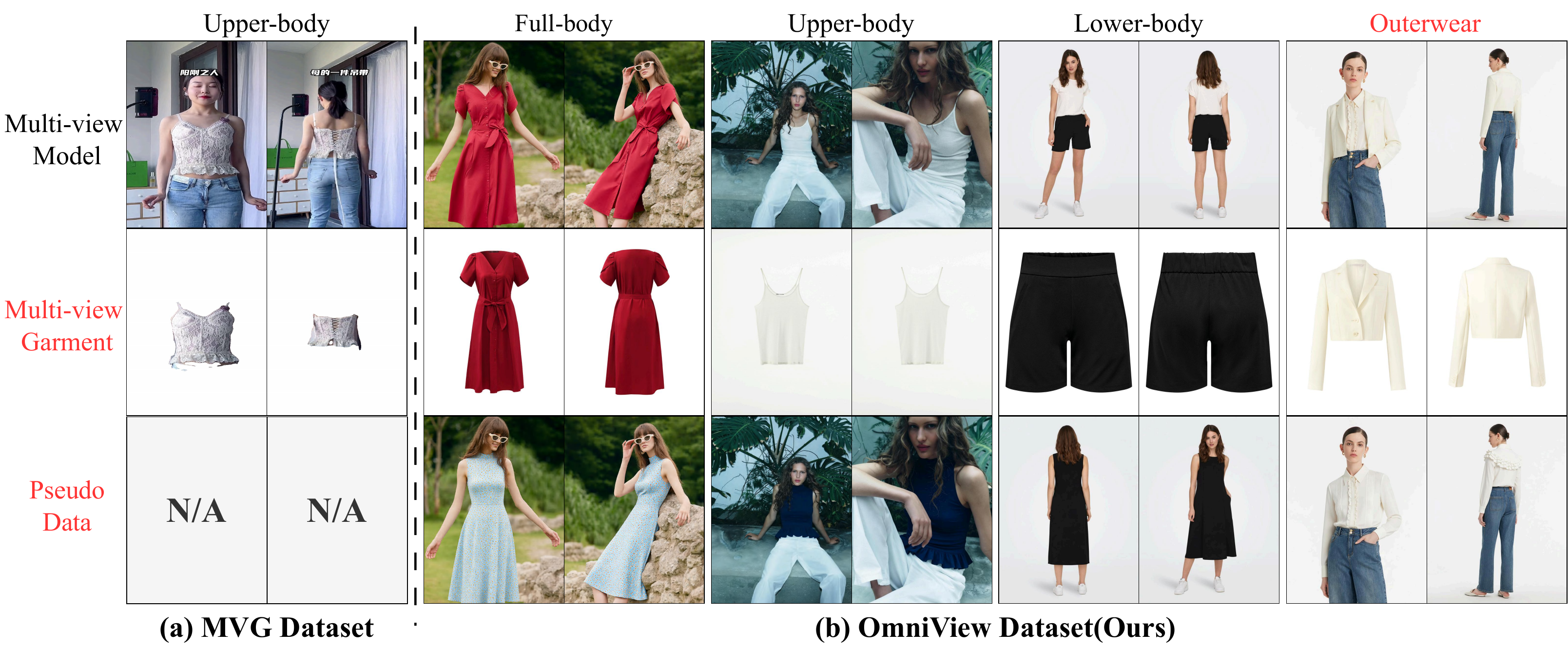}
  \caption{Data samples from MVG (a) and OmniView (b). Compared with MVG, OmniView provides higher-quality garment images, more garment categories, and additional pseudo data.}
  \Description{The figure compares representative samples from MVG in panel (a) and OmniView in panel (b). The OmniView samples include person images from multiple viewpoints, front- and back-view flat garment images, and pseudo data generated through intra-category garment replacement, cross-category garment replacement, and outerwear removal.}
  \label{fig:freeview_examples}
\end{figure*}

\section{OmniView Dataset}
\noindent{\textbf{Overview of OmniView Dataset.}} Existing keyframe-driven VVT methods~\cite{zuo2025dreamvvt} require multi-view try-on images to provide comprehensive garment cues. However, existing multi-view try-on datasets, such as MVG~\cite{wang2025mv}, are limited in scale, garment diversity, and completeness (see Table~\ref{tab:dataset_compare}). To address these limitations, we construct OmniView, a large-scale multi-view try-on dataset containing 6,110 samples. Each sample includes person images from at least two viewpoints, with 88\% containing back-view images. As shown in Fig.~\ref{fig:freeview_examples}, OmniView improves existing datasets in three aspects. First, each sample provides front- and back-view flat garment images for more complete garment references. Second, besides upper-body, lower-body, and full-body categories, OmniView introduces an additional outerwear category. Third, we synthesize pseudo-labeled data for each person image to support cross-category and layered try-on. In Fig.~\ref{fig:freeview_examples}(b), the third and fourth columns show pseudo data for cross-category and layered try-on, respectively.

\noindent{\textbf{Data processing.}} We collect raw data from multiple e-commerce platforms and open-source datasets~\cite{shen2025imagdressing} following their terms of use. To address noise in the raw data, such as inconsistent model identities, we use Qwen3-VL-32B~\cite{bai2025qwen3} to annotate model and garment images, followed by regrouping and manual verification to ensure identity consistency and garment correspondence. We then generate editing instructions with Qwen3-VL-32B and synthesize pseudo data using Qwen-Image-Edit-2511~\cite{wu2025qwen}. For non-outerwear categories, we perform intra- and cross-category garment replacement, while for outerwear, we additionally introduce garment removal for layered try-on. Examples are shown in the third row of Fig.~\ref{fig:freeview_examples}(b), including intra-category replacement (first two columns), cross-category replacement (third column), and garment removal (fourth column). The synthesized data are further reviewed using the same pipeline to ensure quality and consistency.

\begin{figure*}[t]
  \centering
  \includegraphics[width=0.9\textwidth]{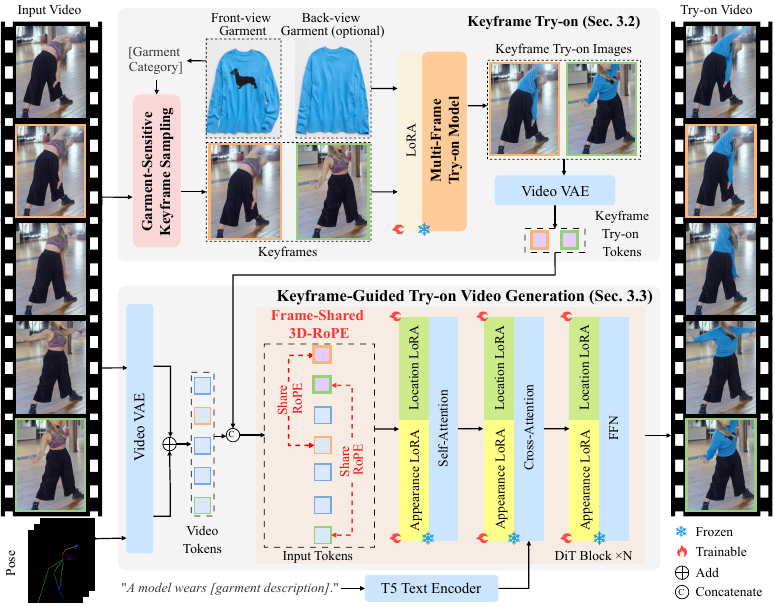}
  \caption{Overview of BooM-VVT. The framework consists of two components: keyframe try-on and keyframe-guided try-on video generation. The first component selects informative keyframes from the input video and generates corresponding try-on images, while the second synthesizes the final try-on video under the guidance of these keyframe try-on results.}
  \Description{The architecture diagram contains two main components. The keyframe try-on component selects garment-sensitive keyframes from the input video and processes the keyframes and garment references with a multi-frame try-on model. The video generation component encodes the input video, pose sequence, and generated keyframe try-on images into tokens. These tokens are processed by a DiT backbone with Frame-Shared 3D-RoPE and decoded to produce the final try-on video.}
  \label{fig:network}
\end{figure*}

\section{Method}
\subsection{Overview}
BooM-VVT is built on the keyframe-driven video generation paradigm and aims to achieve mask-free video virtual try-on, as illustrated in Fig.~\ref{fig:network}. The framework consists of two stages: keyframe try-on (Sec.~3.2) and keyframe-guided try-on video generation (Sec.~3.3). Given an input video $V_{in}$, we first select informative keyframes using the proposed Garment-Sensitive Keyframe Sampling (GSKS) strategy and generate their try-on results with a multi-frame try-on model. Leveraging OmniView, the model supports both single-view and multi-view garment inputs and achieves reliable try-on across diverse viewpoints. The generated keyframe results are then used as appearance conditions for a DiT-based~\cite{peebles2023scalable} video generation model. To facilitate accurate garment appearance transfer, we propose Frame-Shared 3D-RoPE, which aligns keyframe try-on tokens with video-frame tokens through shared positional encodings. We further introduce a multi-stage training strategy to enable mask-free video generation, as detailed in Fig.~\ref{fig:training_framework}.

\subsection{Keyframe Try-on}
\subsubsection{Garment-Sensitive Keyframe Sampling.}
In keyframe-driven VVT, keyframes serve as the primary carrier of target garment information for the final try-on video. An effective keyframe set should satisfy two properties: (i) \textbf{high information capacity}: garment-relevant body regions should be sufficiently visible so that fine-grained garment details can be captured; and (ii) \textbf{complementary viewpoint coverage}: the selected keyframes should cover diverse viewpoints to provide more complete garment information.

To this end, we propose a Garment-Sensitive Keyframe Sampling (GSKS) strategy. To quantify the information capacity of each frame, we first define a garment-relevant limb set $P_c$ and body region $A_c$ according to the target garment category $c$. For example, for upper-body garments, $P_c$ includes limbs such as the shoulders and arms, while $A_c$ corresponds to the upper torso. Given an input video $V_{in}$, GSKS computes an information capacity score $S_{\mathrm{Capacity}}^{c}$ for each frame based on a limb visibility score $S_{p}^{c}$ and a region visibility score $S_{a}^{c}$. Specifically, we use DWPose~\cite{yang2023effective} to detect limbs in each frame, and define $S_{p}^{c}$ as the ratio of the detected limbs belonging to $P_c$ to the total number of limbs in $P_c$:
\begin{equation}
    S_{p}^{c}=\frac{|\hat{P} \cap P_c|}{|P_c|},
\end{equation}
where $\hat{P}$ denotes the set of detected limbs in the current frame.  Meanwhile, SAM~\cite{carion2025sam} estimates the garment-relevant region $A_c$, and the region visibility score $S_a^c$ is computed as its area ratio to the whole frame. The final score is: 
\begin{equation}
    S_{\mathrm{Capacity}}^{c}=(1-\lambda) S_{p}^{c}+\lambda S_{a}^{c}.
\end{equation}
In practice, we set $\lambda=0.2$.

After scoring all frames, we select the frame with the highest $S_{\mathrm{Capacity}}^{c}$ as the primary keyframe and add it to the keyframe set $\mathcal{K}$. We then iteratively select the frame with the largest viewpoint difference from the selected keyframes to expand $\mathcal{K}$, ensuring complementary viewpoint coverage. We measure viewpoint difference as the average cosine distance between the skeletal joint direction vectors of a candidate frame and those of the selected keyframes. To capture garment-relevant appearance variation, we only consider the limbs in $P_c$, avoiding interference from irrelevant body regions. The pseudo-code is provided in the Appendix.

\subsubsection{Multi-Frame Try-on Model}
Given the selected keyframes, we employ a pretrained image try-on model based on the MM-DiT architecture~\cite{esser2024scaling} to generate keyframe try-on images. To improve cross-frame consistency, we adapt it to a multi-frame setting through image concatenation. Specifically, the selected keyframes are horizontally concatenated into a composite image, processed by the model, and then split into individual try-on frames. This single forward pass enables cross-frame information interaction and produces coherent results with consistent garment details. When multi-view garment images are available, we organize them using the same concatenation strategy as garment inputs. We further fine-tune the model on OmniView using lightweight LoRA adapters inserted into the self-attention layers of MM-DiT blocks. During training, both single-view and multi-view garment inputs are used, enabling the model to generate plausible try-on results for challenging viewpoints (e.g., back views) with single-view garments. Multi-view garment references can further improve visual fidelity.

\begin{figure*}[t]
  \centering
  \includegraphics[width=0.9\textwidth]{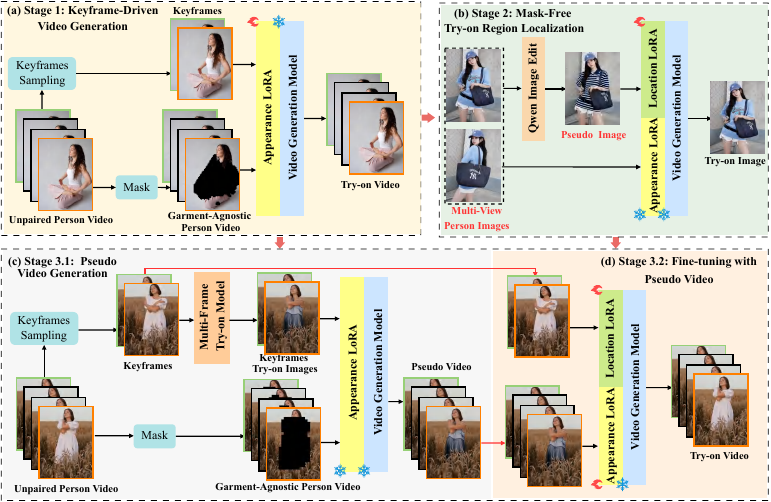}
  \caption{Overview of the proposed multi-stage training strategy for the video generation model in BooM-VVT. (a) Stage 1: The video generation model is trained for keyframe-driven video generation using unpaired videos. (b) Stage 2: The model learns mask-free try-on region localization from image-level pseudo data. (c) Stage 3.1: We use the model trained in Stage~1 to synthesize video-level pseudo data. (d) Stage 3.2: The model is fine-tuned on video-level pseudo data to enable high-quality try-on video generation without masks.}
  \Description{The figure illustrates the four parts of the multi-stage training strategy. Panel (a) trains the video model to reconstruct an original video from a garment-agnostic video and sampled keyframes. Panel (b) uses an edited pseudo image and another image of the same person to learn mask-free garment localization. Panel (c) generates video-level pseudo data from keyframe try-on images and garment-agnostic videos using the Stage 1 model. Panel (d) fine-tunes the Stage 2 model using pseudo videos and keyframes as inputs and the original videos as supervision.}
  \label{fig:training_framework}
\end{figure*}

\subsection{Keyframe-Guided Try-on Video Generation}
Our video generation model is built upon Wan-Animate~\cite{cheng2025wan}, a DiT-based image-to-video framework. As illustrated in Fig.~\ref{fig:network}, given an input video and pose sequence, we encode them into video and pose tokens using a Video VAE. The pose tokens are added to the video tokens for structural guidance. Meanwhile, the keyframe try-on images are encoded into keyframe tokens using the same VAE. These keyframe tokens are concatenated with video tokens along the sequence dimension and fed into the DiT backbone. After denoising, the latent representation is decoded by the VAE decoder to obtain the final try-on video.

\subsubsection{Frame-Shared 3D-RoPE}
Following the standard design of image-to-video generation, existing keyframe-driven VVT methods treat keyframe try-on images and video frames as spatiotemporally independent and assign them different positional encodings. However, this ignores an important prior in keyframe-driven VVT: each keyframe try-on image is naturally aligned with its corresponding target frame. Assigning different positional encodings to such aligned representations increases their relative positional distance, thereby weakening the attention interaction between them and hindering accurate garment appearance transfer.

To address this issue, we propose Frame-Shared 3D-RoPE. Our design builds on the distance-aware property of 3D-RoPE~\cite{su2024roformer}, where tokens with smaller spatiotemporal distances tend to attend more strongly. Specifically, we assign each keyframe try-on token the same 3D positional encoding as its corresponding video-frame token. By sharing positional encodings across aligned keyframe and video-frame tokens, Frame-Shared 3D-RoPE reduces their relative positional distance and strengthens the attention interaction between them, leading to improved garment appearance transfer.

% ================= FIGURE 5: ViViD-S Qualitative =================
\begin{figure*}[t]
    \centering
    \includegraphics[width=0.9\textwidth]{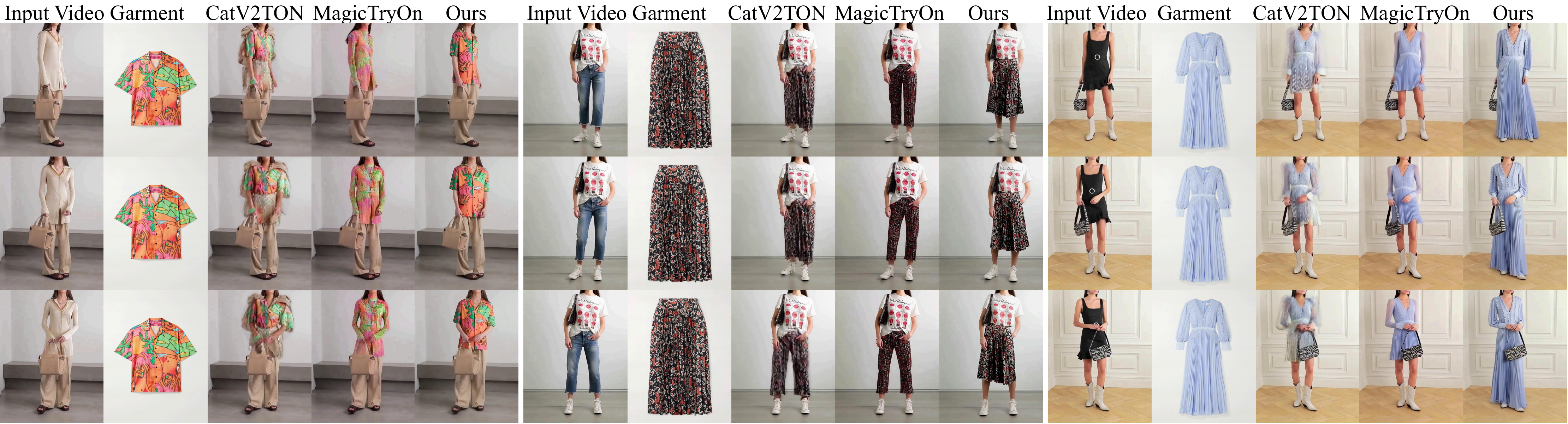}
    \caption{Qualitative comparison on the ViViD-S dataset.}
    \Description{The figure presents several qualitative examples from the ViViD-S dataset. Each example includes source person frames, a target garment reference, and try-on results generated by ViViD, CatV2TON, MagicTryOn, and BooM-VVT. The examples compare garment shape reconstruction, local garment details, person appearance, and temporal consistency across the methods.}
    \label{fig:qual_vivid}
\end{figure*}

\subsubsection{Multi-Stage Training Strategy}
To achieve mask-free video virtual try-on, existing image-based virtual try-on methods typically rely on large-scale pseudo-paired data to learn mask-free try-on region localization. However, directly extending this strategy to videos is impractical due to the high cost of constructing large-scale video-level pseudo data. To address this challenge, we propose a multi-stage training strategy, as illustrated in Fig.~\ref{fig:training_framework}. Specifically, we first train the video model for keyframe-guided generation on large-scale unpaired human-centric videos, then introduce mask-free try-on localization using image-level pseudo data, and finally adapt the model to mask-free video virtual try-on with limited video-level pseudo data.

\noindent\textbf{Stage 1: Training for Keyframe-Driven Video Generation.}
As shown in Fig.~\ref{fig:training_framework}(a), in the first stage, we follow the standard keyframe-driven training paradigm and train the video generation model on large-scale unpaired human-centric videos. Specifically, for each input video, we first randomly sample several frames as keyframe references. We then use garment masks to remove the original garments from the video, obtaining a garment-agnostic input. Given this garment-agnostic video and the sampled keyframes, the model is trained to reconstruct the original video. Since try-on regions are explicitly provided by masks, the model focuses on learning garment appearance transfer and temporal coherence.

\noindent\textbf{Stage 2: Training for Mask-Free Try-on Localization.}
As illustrated in Fig.~\ref{fig:training_framework}(b), the second stage equips the video model with mask-free try-on localization capability. Instead of constructing costly video-level pseudo data, we leverage readily available image-level pseudo pairs for supervision. Specifically, we build a large-scale multi-view person dataset by sampling frames from videos and incorporating existing multi-view person datasets, where each sample contains two images of the same person under different poses. For each sample, one image is selected as the target image, whose garment is edited by Qwen-Image-Edit-2511 to generate a pseudo image, while the other image serves as the keyframe reference. The model is then trained to reconstruct the target image without explicit masks, forcing it to learn mask-free garment replacement localization. To preserve the priors learned in Stage~1, we freeze Appearance LoRA and introduce a separate Location LoRA.

\noindent\textbf{Stage 3: Training for High-Quality Mask-Free Video Try-on.}
Although Stage~2 enables mask-free try-on, image-level supervision lacks temporal constraints, resulting in limited video consistency. Therefore, we introduce a final fine-tuning stage with video-level pseudo data. As shown in Fig.~\ref{fig:training_framework}(c,d), Stage~3 consists of two sub-stages: Stage~3.1 for pseudo video generation and Stage~3.2 for fine-tuning with pseudo videos. In Stage~3.1, we use the Stage~1 model to construct video-level pseudo data from unpaired videos. Specifically, we generate keyframe try-on images using the multi-frame try-on model and construct garment-agnostic videos by removing original garments with garment masks. Conditioned on these inputs, the Stage~1 model synthesizes pseudo videos. In Stage~3.2, we fine-tune the Stage~2 model using pseudo videos and corresponding keyframes, with original unpaired videos as supervision. This stage combines the appearance transfer capability from Stage~1 with the mask-free localization capability from Stage~2, producing high-quality try-on videos with improved temporal coherence.

\begin{table*}[t]
    \centering
    \caption{Quantitative results on the ViViD-S dataset.}
    \label{tab:quant_vivid}
    \small
    \begin{tabular}{l c cccc ccc}
            \toprule
            \multirow{2}{*}{Method} & \multirow{2}{*}{Mask} & \multicolumn{4}{c}{Unpaired Setting} & \multicolumn{3}{c}{Paired Setting} \\
            \cmidrule(lr){3-6} \cmidrule(lr){7-9}
            & & FVD$^u \downarrow$ & ABC $\uparrow$ & GC $\uparrow$ & OVQ $\uparrow$ & FVD$^p \downarrow$ & SSIM $\uparrow$ & LPIPS $\downarrow$ \\
            \midrule
            ViViD & \cmark & 164.87 & 3.03 & 2.26 & 2.24 & 139.10 & 0.842 & 0.078 \\
            CatV2TON & \cmark & 128.53 & 3.53 & 2.07 & 1.98 & 120.38 & 0.858 & 0.077 \\
            MagicTryOn & \cmark & 110.66 & 4.15 & 2.38 & 3.12 & \phantom{0}86.31 & 0.869 & 0.068 \\
            \midrule
             \textbf{BooM-VVT (Stage 1)} & \cmark & 106.25 & 4.38 & 3.36 & 3.57 & \phantom{0}88.19 & 0.872 & 0.065 \\
            \textbf{BooM-VVT} & \xmark & \textbf{101.92} & \textbf{4.61} & \textbf{3.95} & \textbf{4.21} & \textbf{\phantom{0}79.12} & \textbf{0.891} & \textbf{0.061} \\
            \bottomrule
    \end{tabular}
\end{table*}

\section{Experiments}
\subsection{Experimental Setup}
\noindent{\textbf{Training Datasets.}}
For the multi-frame try-on model, we use OmniView, MVG~\cite{wang2025mv}, and ViViD~\cite{fang2024vivid} datasets. From each ViViD video, we sample two frames as multi-view person images to construct training pairs. In total,  we obtain approximately 22K pairs, each containing two person images in different poses and their corresponding garment images. The video generation model uses stage-specific data. For Stage~1, we collect 20K human-centric videos from public datasets~\cite{wang2024humanvid,li2025openhumanvid,fang2024vivid} and online sources. For Stage~2, we sample two frames from each Stage~1 video and combine them with OmniView, yielding 30K multi-view pairs for mask-free try-on localization. For Stage~3, we select 2,500 Stage~1 videos and synthesize their corresponding pseudo videos for final fine-tuning.

\begin{figure*}[t]
    \centering
    \includegraphics[width=0.9\textwidth]{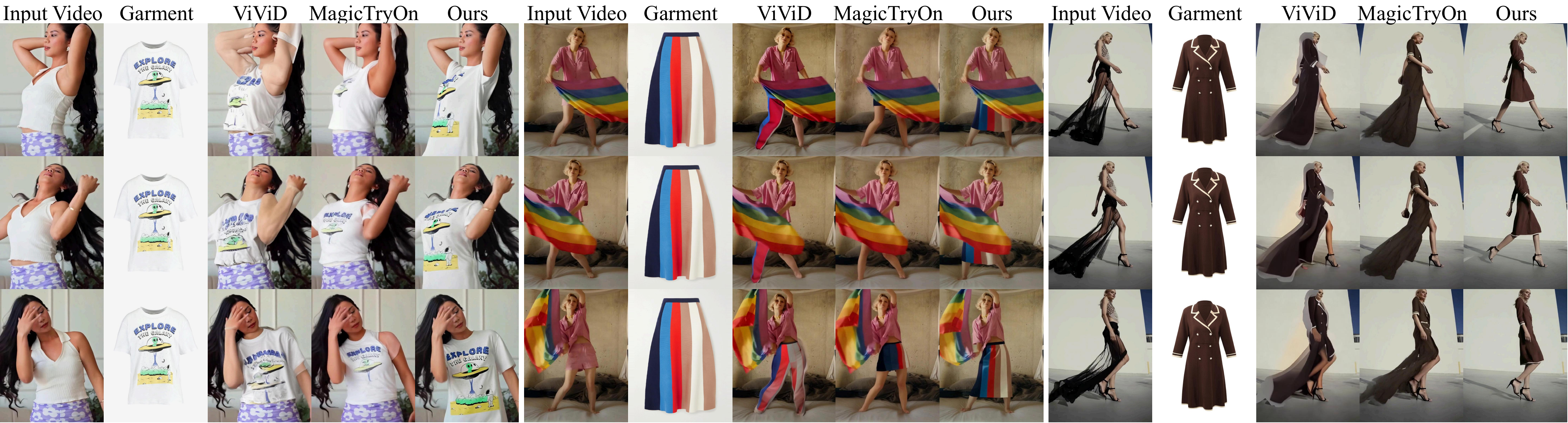} % 请替换文件名
    \caption{Qualitative comparison on the WildVVT benchmark.}
    \Description{The figure presents qualitative comparisons on the WildVVT benchmark. The examples include source videos with turning motions, large pose changes, severe occlusions, and complex real-world backgrounds, together with target garment references and results from ViViD, CatV2TON, MagicTryOn, and BooM-VVT. The generated frames illustrate differences in garment structure, fine-grained details, appearance consistency, and temporal coherence.}
    \label{fig:qual_wild}
\end{figure*}

\noindent{\textbf{Evaluation Datasets.}}
For indoor evaluation, we use ViViD-S, which contains 180 samples constructed following the CatV2TON protocol. In addition, we create an in-the-wild benchmark, named WildVVT, consisting of 100 samples that cover diverse human motions, scene conditions, and garment categories.

\noindent{\textbf{Implementation Details.}}
For keyframe try-on, we adopt Qwen-Image-Edit-2511 as the Multi-Frame Try-on model and train a rank-64 LoRA for 20K steps. For video generation, we use Wan-Animate as the DiT backbone. We train a rank-128 Appearance LoRA for 10K steps in Stage~1, freeze it and train a rank-64 Location LoRA for 20K steps in Stage~2, and jointly fine-tune both LoRAs for 5K steps in Stage~3. Unless otherwise specified, all models are trained with AdamW on 4 NVIDIA RTX Pro 6000 GPUs, using a learning rate of \(1\times10^{-4}\) and a batch size of 1. During inference, following DreamVVT, we sample two keyframes per video to balance efficiency and generation quality. We combine the Multi-Frame Try-on model with Qwen-Image-Edit-Lightning~\cite{qwen_image_edit_lightning_github}, an acceleration LoRA for Qwen-Image-Edit-2511, and use 8 inference steps. The same acceleration LoRA is used for pseudo-image generation during data construction. For video generation, we use 20 inference steps with a guidance scale of 1.

\noindent\textbf{Evaluation Metrics.}
Following prior work, we evaluate generated try-on videos using FVD~\cite{unterthiner2018towards}, SSIM~\cite{wang2004image}, and LPIPS~\cite{zhang2018unreasonable}. We further employ Gemini 3.1 Pro as a judge, scoring each result on a 0--5 scale along three dimensions: \textbf{Appearance and Background Consistency (ABC)}, measuring the preservation of identity, background, and other non-try-on regions; \textbf{Garment Consistency (GC)}, measuring the faithfulness of the generated garment to the reference; and \textbf{Overall Video Quality (OVQ)}, measuring realism and temporal coherence.

\subsection{Qualitative Comparison}
We qualitatively compare BooM-VVT with state-of-the-art VVT methods, including ViViD~\cite{fang2024vivid}, CatV2TON~\cite{chong2025catv2ton}, and MagicTryOn~\cite{li2025magictryon}. As shown in Figs.~\ref{fig:qual_vivid} and~\ref{fig:qual_wild}, BooM-VVT produces more realistic and temporally coherent results on both ViViD-S and the more challenging WildVVT benchmark. CatV2TON and MagicTryOn often exhibit artifacts caused by their mask-based designs. As illustrated in Fig.~\ref{fig:qual_vivid}, when the source and target garments differ substantially in shape, they remain constrained by the original garment structure and fail to reconstruct the target garment accurately. Moreover, as shown in Fig.~\ref{fig:qual_wild}, under challenging turning motions and severe occlusions, ViViD and MagicTryOn often fail to preserve plausible garment structures and fine-grained details. In contrast, BooM-VVT achieves better garment fidelity and spatiotemporal consistency across diverse real-world scenarios. Additional qualitative results are provided in the Appendix.

\begin{table*}[t]
    \centering
    \caption{Quantitative results on the WildVVT benchmark. The inference time and GPU memory usage are measured on a single NVIDIA A800 when generating a 65-frame video at a resolution of $624\times832$.}
    \label{tab:quant_wild}
    \small % 字体大小
    \begin{tabular}{l c c c c c c c}
            \toprule
            Method & Mask & FVD $\downarrow$ & ABC $\uparrow$ & GC $\uparrow$ & OVQ $\uparrow$ & \makecell{GPU\\Memory} & \makecell{Inference\\Time} \\
            \midrule
             ViViD & \cmark & 301.80 & 2.13 & 1.55 & 1.57 & 55.8G & \textbf{142s} \\
             CatV2TON & \cmark & 350.72 & 1.39 & 1.28 & 1.23 & \textbf{28.9G} & 158s \\
             MagicTryOn & \cmark & 250.55 & 2.12 & 1.84 & 1.65 & 63.5G & 342s \\
            \midrule
            \textbf{BooM-VVT (Stage 1)} & \cmark & 241.72 & 3.96 & 3.07 & 3.26 & 57.9G & 281s \\
            \textbf{BooM-VVT} & \xmark & \textbf{219.76} & \textbf{4.33} & \textbf{3.72} & \textbf{3.94} & 57.9G & 281s \\
            \bottomrule
    \end{tabular}
\end{table*}

\begin{table}[t]
    \centering
    \caption{Ablation study on the WildVVT benchmark.}
    \label{tab:ablation}
    \small
    \begin{tabular}{l c c c c}
        \toprule
        Method & FVD $\downarrow$ & ABC $\uparrow$ & GC $\uparrow$ & OVQ $\uparrow$ \\
        \midrule
        w/o GSKS & 227.88 & 4.21 & 3.42 & 3.85 \\
        w/o Frame-Shared 3D-RoPE & 224.15 & \textbf{4.34} & 3.27 & 3.81 \\
        w/o Stage~1 & 242.61 & 3.91 & 3.26 & 3.66 \\
        w/o Stage~2 & 237.89 & 4.10 & 3.37 & 3.75 \\
        \midrule
        \textbf{Full Model} & \textbf{219.76} & 4.33 & \textbf{3.72} & \textbf{3.94} \\
        \bottomrule
    \end{tabular}
\end{table}

\begin{figure*}[t]
    \centering
    \includegraphics[width=\textwidth]{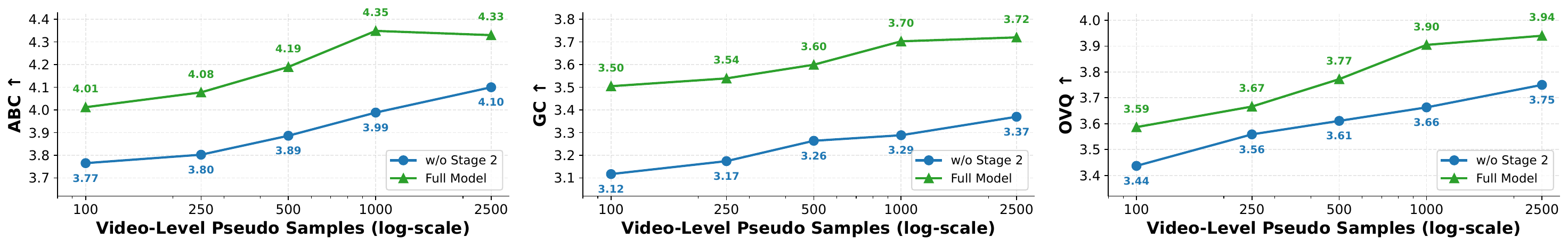}
    \caption{Effectiveness of Stage~2 under different amounts of video-level pseudo data on the WildVVT benchmark.}
    \Description{The plot compares the full model with the variant without Stage 2 under different numbers of video-level pseudo samples. The horizontal axis represents the amount of video-level pseudo data, and the performance curves show that the full model consistently performs better across all data scales. The full model using 500 pseudo videos reaches a performance level comparable to the variant without Stage 2 using 2,500 pseudo videos.}
\label{fig:ablation_stage2_curve}
\end{figure*}

\begin{figure}[ht]
    \centering
    \includegraphics[width=0.9\columnwidth]{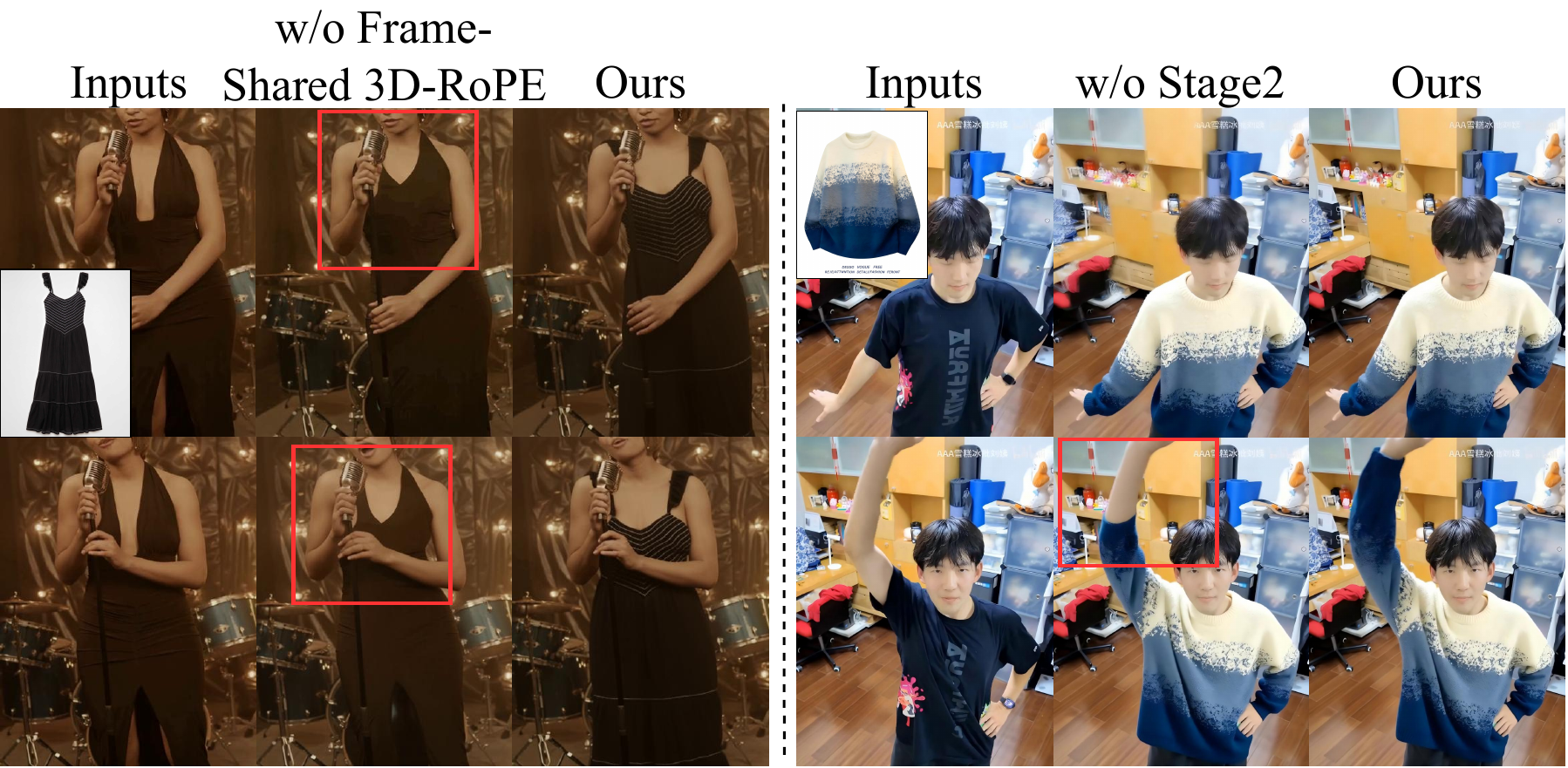}
    \caption{Qualitative results of ablation study on the WildVVT benchmark.}
    \Description{The figure contains two qualitative ablation comparisons on WildVVT. The left comparison shows the full model and the variant without Frame-Shared 3D-RoPE, focusing on the reconstruction of shoulder straps and white stripes on the chest. The right comparison shows the full model and the variant without Stage 2, highlighting differences in try-on region localization and generated garment structure.}
    \label{fig:ablation_rope_stage}
\end{figure}

\subsection{Quantitative Comparison}
Quantitative results are reported in Table~\ref{tab:quant_vivid} and Table~\ref{tab:quant_wild}. Since public implementations of existing keyframe-driven VVT methods are unavailable, direct comparison with them is infeasible. 
Therefore, we use BooM-VVT (Stage~1) as a baseline for the keyframe-driven setting. Specifically, BooM-VVT (Stage~1) is a variant of our method whose video model is trained only in Stage~1. The results show that the keyframe-driven methods, BooM-VVT (Stage~1) and BooM-VVT, consistently outperform the non-keyframe-driven baselines on both ViViD-S and WildVVT, demonstrating the effectiveness of the keyframe-driven paradigm. However, mask dependence remains a major limitation of existing baselines. 
When source and target garments differ substantially in shape, inaccurate masks hinder garment transfer and reduce GC scores, consistent with the qualitative results in Fig.~\ref{fig:qual_vivid}. The limitation becomes more pronounced on the more challenging WildVVT benchmark, where large pose variations and severe occlusions make it particularly difficult to obtain accurate masks. As a result, existing methods often fail to maintain appearance consistency and exhibit temporal jittering and flickering, leading to lower ABC and OVQ scores. By contrast, BooM-VVT remains robust in unconstrained real-world scenarios and achieves the best performance across all quality metrics. Table~\ref{tab:quant_wild} further reports inference time and GPU memory usage. Compared with MagicTryOn, BooM-VVT achieves substantially better performance at comparable inference cost.

\subsection{Ablation Study}
We conduct ablation studies on the WildVVT benchmark to evaluate the effectiveness of the key components in BooM-VVT, covering both architectural design and training strategy.

\noindent\textbf{On the model architecture.}
We first examine two key components, Garment-Sensitive Keyframe Sampling (GSKS) and Frame-Shared 3D-RoPE. As shown in Table~\ref{tab:ablation}, removing either component degrades performance, particularly Garment Consistency (GC). For the \emph{w/o GSKS} variant, we follow DreamVVT and select keyframes mainly based on global pose differences. For the \emph{w/o Frame-Shared 3D-RoPE} variant, keyframe try-on images and their corresponding target video frames are assigned different positional encodings, which weakens their attention interaction and hinders garment appearance transfer. Although this variant achieves a slightly higher ABC score, its GC and OVQ decrease while FVD increases, indicating that it preserves non-try-on regions at the expense of garment transfer quality. Fig.~\ref{fig:ablation_rope_stage} further shows that this variant reconstructs incorrect shoulder straps and misses the white chest stripes.

\noindent\textbf{On the training strategy.}
We further analyze the multi-stage training strategy, with results reported in Table~\ref{tab:ablation} and Fig.~\ref{fig:ablation_stage2_curve}.
The \emph{w/o Stage~1} variant skips Stage~1 training for the Appearance LoRA, while the \emph{w/o Stage~2} variant skips Stage~2 training for the Location LoRA. To ensure a fair comparison in data construction cost, we train \emph{w/o Stage~2} with 400 additional video-level pseudo samples, whose construction cost matches that of the 30K image-level pseudo samples used in Stage~2.

As shown in Table~\ref{tab:ablation}, removing Stage~1 causes the largest performance drop, confirming its importance for high-quality keyframe-driven try-on video generation. Removing Stage~2 also clearly degrades performance, demonstrating its role in accurate mask-free try-on localization. This is further illustrated in the right panel of Fig.~\ref{fig:ablation_rope_stage}, where \emph{w/o Stage~2} incorrectly localizes the try-on region and produces erroneous garment structures. Fig.~\ref{fig:ablation_stage2_curve} further shows that Stage~2 substantially improves fine-tuning efficiency at comparable data construction costs. Across all pseudo-video scales, the full model consistently outperforms the \emph{w/o Stage~2} variant. Notably, with only 500 video-level pseudo samples, the full model achieves performance comparable to that of \emph{w/o Stage~2} trained with 2,500 samples. We attribute this advantage to the broad coverage of garment types, poses, and real-world scenes in the large-scale image-level pseudo data, which improves generalization to unconstrained scenarios.

\section{Conclusion}
In this paper, we present BooM-VVT, a novel mask-free video virtual try-on framework built on the keyframe-driven paradigm. To achieve mask-free video virtual try-on, we introduce a multi-stage training strategy that leverages readily available image-level pseudo data to learn mask-free try-on region localization, substantially reducing the reliance on costly video-level pseudo data. To improve garment consistency, we further introduce Garment-Sensitive Keyframe Sampling (GSKS) and Frame-Shared 3D-RoPE. GSKS selects informative keyframes based on garment-relevant body regions, while Frame-Shared 3D-RoPE improves garment appearance transfer by explicitly establishing spatiotemporal correspondences between keyframes and their corresponding target video frames. In addition, we construct OmniView, a large-scale multi-view try-on dataset that provides more complete garment references and broader garment category coverage, enabling reliable try-on generation under challenging viewpoints and diverse tasks. Extensive experiments demonstrate that BooM-VVT outperforms existing methods in temporal consistency and garment fidelity under unconstrained real-world scenarios.

\section{Limitations}
Although BooM-VVT demonstrates strong robustness in challenging
real-world scenarios, GSKS still relies on off-the-shelf DWPose
and SAM. Extremely severe occlusions or inaccurate detections may
therefore lead to suboptimal keyframe selection. In addition,
extreme illumination changes may affect mask-free localization.
Improving robustness to imperfect visual priors and reducing
inference cost remain important directions for future work.

%%
%% The acknowledgments section is defined using the "acks" environment
%% (and NOT an unnumbered section). This ensures the proper
%% identification of the section in the article metadata, and the
%% consistent spelling of the heading.
\begin{acks}
This work was partially supported by the 2024 Jiangsu
Province Frontier Technology Research and Development Project (Grant No. BF2024071) and the Jiangsu Provincial Science and Technology Major Project (Grant No.
BG2024042).
\end{acks}

%%
%% The next two lines define the bibliography style to be used, and
%% the bibliography file.
\bibliographystyle{ACM-Reference-Format}
\balance
\bibliography{sample-base}

\clearpage
\appendix
\section{Related Work}
\subsection{Image Virtual Try-On}
Image virtual try-on aims to synthesize a realistic image of a target person wearing a reference garment while preserving identity and scene context. Early methods~\cite{han2018viton,wang2018toward,xie2023gp} mainly followed a warping-and-fusion pipeline, where the garment was first aligned to the target body via TPS or learned flow and then blended into the person image by a GAN-based generator. Although effective in controlled settings, these methods often suffer from misalignment artifacts and limited robustness under complex poses, severe occlusions, and unconstrained scenes.

Recent advances in diffusion models~\cite{ho2020denoising,nichol2021improved,rombach2022high} have substantially improved synthesis realism and garment fidelity in image virtual try-on. Representative methods~\cite{choi2024improving,xu2025ootdiffusion,jiang2024fitdit,chong2024catvton} build on latent diffusion model~\cite{rombach2022high} and further enhance garment preservation through reference-feature injection or simple input concatenation. Beyond standard frontal-view try-on, MV-VTON~\cite{wang2025mv} extends image virtual try-on to multi-view settings by using front- and back-view garment images, and constructs the MVG dataset for this task. However, most existing image try-on methods still rely on dense preprocessing, such as human parsing and garment masks, which limits robustness and practical usability in real-world scenarios. To alleviate this limitation, more recent studies~\cite{guo2025any2anytryon,zhang2026unifit,feng2025omnitry} explore mask-free paradigms, where the model implicitly learns try-on region localization from large-scale pseudo-paired data. As a result, these methods exhibit improved flexibility and generalization. However, their effectiveness has been demonstrated primarily in image synthesis, and extending mask-free paradigms to temporally coherent video generation remains challenging.

\subsection{Video Virtual Try-On}
Compared with image virtual try-on, video virtual try-on (VVT) additionally requires temporal coherence across frames. Early VVT methods typically extend image-based pipelines to the video setting by introducing dedicated temporal modules. For example, FW-GAN~\cite{dong2019fw} extends image-based try-on to videos through optical-flow-guided warping and fusion, while MV-TON~\cite{zhong2021mv} improves temporal consistency with a memory refinement module that propagates information from previously generated frames. Although these methods can produce plausible try-on results, their ability to maintain temporal coherence is often limited by the generative capacity of GAN-based architectures.

With the success of diffusion models in video generation, recent VVT methods have made significant progress. ViViD~\cite{fang2024vivid} adapts image diffusion models to video try-on by introducing temporal modeling modules and a dedicated garment encoder. CatV2TON~\cite{chong2025catv2ton} adopts a video DiT architecture to unify image and video try-on within a single model through temporal concatenation of garment and person inputs. MagicTryOn~\cite{li2025magictryon} introduces a garment-preservation strategy that decomposes garment cues into semantic, structural, and appearance streams, and further incorporates garment-aware spatiotemporal RoPE to reduce temporal jitter. Despite these advances, most existing VVT methods still rely heavily on scarce paired video try-on data, which limits their generalization to unconstrained real-world scenarios.

To reduce this dependence, DreamVVT~\cite{zuo2025dreamvvt} shows that a keyframe-driven, stage-wise design can better exploit readily available unpaired videos to improve video-model generalization. KeyTailor~\cite{he2025devil} introduces a keyframe-driven detail injection strategy to enrich garment dynamics and background integrity. Nevertheless, these methods still depend on explicit masks, which restrict garment fidelity and temporal coherence in challenging real-world scenarios. PEMF-VTO~\cite{chang2025pemf} explores mask-free video virtual try-on by introducing point-enhanced guidance for spatial garment transfer and temporal coherence. However, its learning paradigm still relies on large-scale video-level pseudo data. In contrast, our method achieves mask-free VVT under the keyframe-driven paradigm using more accessible image-level pseudo data, substantially reducing the dependence on video-level pseudo data. We further improve garment consistency in keyframe-driven try-on video generation through garment-sensitive keyframe sampling and explicit alignment between keyframes and target video frames.

\section{Preliminary}
\subsection{Virtual Try-On Paradigms}
\subsubsection{\textbf{Mask-based and Mask-free Try-on.}}
Virtual try-on methods can be broadly categorized into mask-based and mask-free paradigms. Mask-based methods formulate try-on as a conditional inpainting problem, where an agnostic mask is used to specify the editable clothing region. Given a person image $I$, a garment image $I_g$, and a mask $M$, the try-on result $\hat{I}$ can be written as
\begin{equation}
\hat{I} = G(I, I_g, M),
\end{equation}
where $G$ denotes the try-on model. This design provides explicit spatial controllability, but its performance heavily depends on mask quality. In complex scenarios, inaccurate masks may destroy original appearance cues and introduce artifacts. By contrast, mask-free methods do not rely on explicit masks and instead require the model to implicitly localize the try-on region from data. The try-on process can be formulated as
\begin{equation}
\hat{I} = G(I, I_g).
\end{equation}

\subsubsection{\textbf{Keyframe-Driven Video Try-on.}}
Existing video virtual try-on methods are often trained in an end-to-end manner, directly generating a try-on video from garment and video conditions. Given an input video $V=\{I_t\}_{t=1}^{T}$ and a garment image $I_g$, the generated try-on video $\hat{V}$ can be written as
\begin{equation}
\hat{V} = G_{\mathrm{vvt}}(V, I_g).
\end{equation}
Although conceptually simple, this design heavily depends on scarce paired garment--video data, making it difficult to preserve garment fidelity and temporal consistency in unconstrained scenarios. To address this limitation, recent methods adopt a keyframe-driven paradigm, which decomposes VVT into two stages: keyframe try-on and keyframe-guided video generation. Specifically, given sampled keyframes $V_k=\{I_{t_i}\}_{i=1}^{K}$, the keyframe try-on results are first generated as
\begin{equation}
\hat{V}_k = G_{\mathrm{img}}(V_k, I_g),
\end{equation}
and the final try-on video is then synthesized as
\begin{equation}
\hat{V} = G_{\mathrm{vid}}(V, \hat{V}_k).
\end{equation}
This design makes it easier to exploit powerful image try-on models and pretrained video generation backbones, while also enabling training with more readily available unpaired videos. However, existing keyframe-driven methods still mostly rely on explicit masks during try-on generation, which limits their robustness in challenging real-world scenarios.

\subsection{Diffusion Transformers}
Diffusion Transformers (DiTs) have emerged as powerful backbones for image and video generation due to their strong scalability and ability to model long-range dependencies. 

For image generation, recent representative models such as Qwen-Image and FLUX adopt a multimodal diffusion transformer (MM-DiT) architecture. In general, such models consist of a variational autoencoder (VAE), a text encoder, and an MM-DiT backbone. The input image is first mapped into a latent space by the VAE, while the text prompt is encoded into semantic embeddings by the text encoder. The MM-DiT backbone is composed of stacked Transformer blocks, which enable effective interaction across different modalities through self-attention, thereby supporting high-quality image generation and editing. In our framework, the multi-frame try-on model is built upon Qwen-Image-Edit, whose backbone also follows the MM-DiT architecture.

For video generation, recent mainstream models such as Wan adopt a DiT-based latent video generation architecture. A typical video DiT consists of a VAE, a text encoder, and a DiT backbone that includes patchifying and unpatchifying modules together with multiple Transformer blocks. By operating on spatiotemporal latent tokens, the model can jointly capture temporal continuity and appearance consistency across frames. To better incorporate textual instructions in long-context scenarios, such architectures often combine self-attention and cross-attention to inject conditioning information. In our framework, the video generation model is based on Wan-Animate, which is also built on the video DiT architecture.

Recent DiT-based generative models are often trained with \emph{flow matching}. Let $x_1$ denote the latent representation of a target sample and $x_0 \sim \mathcal{N}(0, I)$ denote a Gaussian noise sample. Given a sampled time step $t \in [0,1]$, the interpolated latent is defined as
\begin{equation}
x_t = t x_1 + (1-t)x_0.
\end{equation}
The corresponding target velocity field is
\begin{equation}
v_t = \frac{d x_t}{dt} = x_1 - x_0.
\end{equation}
Flow matching trains the network to predict this velocity field under the given conditions, enabling the model to learn a continuous transport from noise to data. In our framework, both the multi-frame try-on model and the video generation model are trained with \emph{flow matching}.

\section{Method Details}
\subsection{Keyframe Sampling Algorithm}

\begin{figure*}[t]
    \centering
    \includegraphics[width=0.85\linewidth]{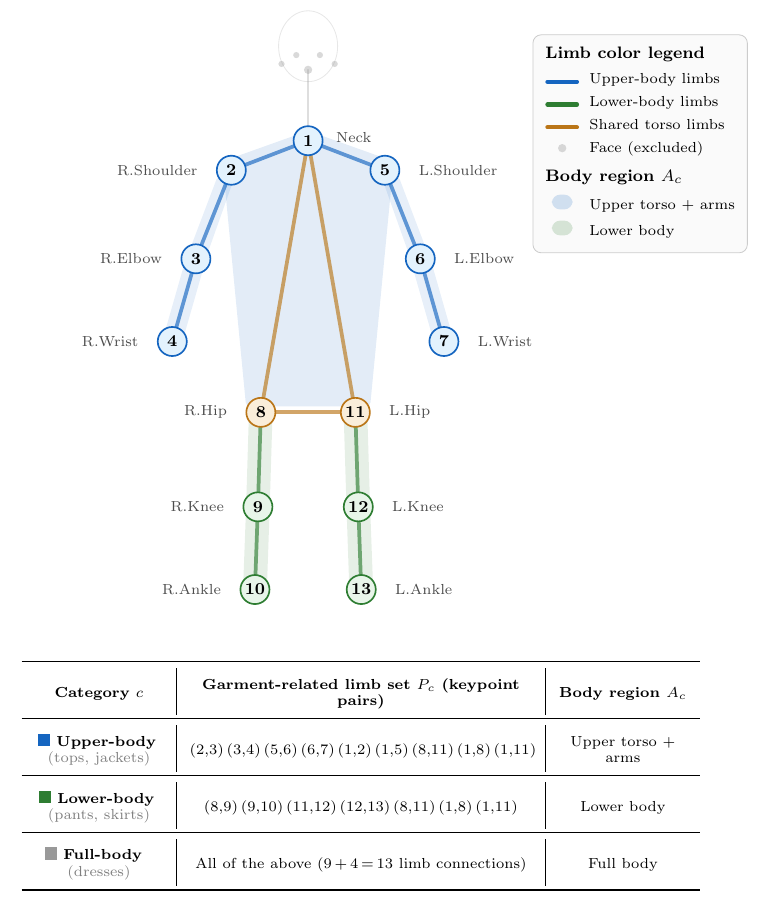}
    \caption{DWPose 18-point skeleton and the definitions of the garment-related limb set $P_c$ and body region $A_c$ for each garment category used in GSKS.}
    \label{fig:skeleton}
\end{figure*}

Fig.~\ref{fig:skeleton} shows the DWPose skeleton used in GSKS, together with the definitions of the garment-related limb set $P_c$ and body region $A_c$ for each garment category. Face-related keypoints (indices 0 and 14--17) are excluded from $P_c$ for all garment categories, as they are irrelevant to garment appearance. As described in Sec.~3.2.1 of the main paper, GSKS selects keyframes according to two criteria: \textit{information capacity} and \textit{complementary viewpoint coverage}. Specifically, for upper-body garments, $P_c$ includes the shoulder and arm limbs together with torso connections; for lower-body garments, $P_c$ includes the hip and leg limbs together with torso connections; and for full-body garments, $P_c$ includes all body limbs. In all experiments, we use DWPose~\cite{yang2023effective} for limb detection and SAM~\cite{carion2025sam} for body region estimation. Unless otherwise specified, we set the number of keyframes to $K=2$, the weighting factor to $\lambda=0.2$, and the minimum frame interval between selected keyframes to $\Delta_{\min}=4$ frames.

To facilitate reproducibility, we provide the complete pseudocode of GSKS in Algorithm~\ref{alg:gsks}. In the complementary viewpoint selection stage, the viewpoint difference between a candidate frame and the selected keyframes is measured using the average cosine distance of garment-related limb direction vectors. Missing limbs are treated as maximally similar to avoid overestimating viewpoint differences caused by unreliable detections under severe occlusion.

\begin{algorithm}[t]
\small
\caption{Garment-Sensitive Keyframe Sampling (GSKS)}
\label{alg:gsks}

\KwIn{
    $V_{\text{in}}$: input video ($N$ frames)\;
    $c$: target garment category\;
    $K$: number of keyframes\;
    $\lambda = 0.2$: weight for region visibility\;
    $\Delta_{\min}=4$: minimum frame interval between selected keyframes\;
}
\KwOut{
    $\mathcal{K}$: selected keyframe set\;
}

Determine garment-related limb set $P_c$ and body region $A_c$ according to $c$\;

\BlankLine
\tcc{Phase 1: Per-frame information-capacity scoring}
\For{$i = 0$ \KwTo $N{-}1$}{
    Detect limbs in $V_{\text{in}}[i]$ using DWPose\;
    
    \ForEach{limb $l=(p_1,p_2)\in P_c$}{
        \eIf{$l$ is detected}{
            $\mathbf{d}_i^l \leftarrow \mathrm{normalize}(p_2-p_1)$; \quad $\mathrm{valid}_i^l \leftarrow \mathrm{True}$\;
        }{
            $\mathbf{d}_i^l \leftarrow \mathbf{0}$; \quad $\mathrm{valid}_i^l \leftarrow \mathrm{False}$\;
        }
    }
    
    Compute limb visibility
    $S_p^c(i)=\frac{|\hat{P}\cap P_c|}{|P_c|}$, where $\hat{P}$ is the detected limb set in frame $i$\;
    
    Estimate garment-related body region $A_c$ using SAM and compute
    $S_a^c(i)=\frac{\mathrm{area}(A_c)}{\mathrm{area}(V_{\text{in}}[i])}$\;
    
    Compute information capacity
    $S_{\mathrm{Capacity}}^c(i)=(1-\lambda)S_p^c(i)+\lambda S_a^c(i)$\;
}

\BlankLine
\tcc{Phase 2: Anchor keyframe selection}
$i^* \leftarrow \arg\max_i S_{\mathrm{Capacity}}^c(i)$\;
$\mathcal{K} \leftarrow \{V_{\text{in}}[i^*]\}$; \quad $\mathcal{I}_{\mathrm{sel}} \leftarrow \{i^*\}$\;

\BlankLine
\tcc{Phase 3: Iterative complementary selection}
\For{$r = 1$ \KwTo $K{-}1$}{
    $i_{\mathrm{best}} \leftarrow \mathrm{None}$; \quad $\mathrm{score}_{\mathrm{best}} \leftarrow -\infty$\;
    
    \ForEach{candidate frame $i \notin \mathcal{I}_{\mathrm{sel}}$}{
        \If{$\exists\, j \in \mathcal{I}_{\mathrm{sel}} : |i-j| < \Delta_{\min}$}{
            \textbf{continue}\;
        }
        
        \ForEach{$j \in \mathcal{I}_{\mathrm{sel}}$}{
            $\mathrm{sim}(i,j)\leftarrow \frac{1}{|P_c|}\sum_{l\in P_c}
            \begin{cases}
                \cos(\mathbf{d}_i^l,\mathbf{d}_j^l), & \text{if } \mathrm{valid}_i^l \wedge \mathrm{valid}_j^l \\
                1.0, & \text{otherwise}
            \end{cases}$\;
        }
        
        $\mathrm{diff}(i)\leftarrow 1-\mathrm{mean}_{j\in\mathcal{I}_{\mathrm{sel}}}(\mathrm{sim}(i,j))$\;
        
        \If{$(\mathrm{diff}(i),\, S_{\mathrm{Capacity}}^c(i)) > \mathrm{score}_{\mathrm{best}}$}{
            $i_{\mathrm{best}} \leftarrow i$; \quad $\mathrm{score}_{\mathrm{best}} \leftarrow (\mathrm{diff}(i),\, S_{\mathrm{Capacity}}^c(i))$\;
        }
    }
    
    \eIf{$i_{\mathrm{best}} \neq \mathrm{None}$}{
        $\mathcal{K} \leftarrow \mathcal{K} \cup \{V_{\text{in}}[i_{\mathrm{best}}]\}$\;
        $\mathcal{I}_{\mathrm{sel}} \leftarrow \mathcal{I}_{\mathrm{sel}} \cup \{i_{\mathrm{best}}\}$\;
    }{
        \textbf{break}\;
    }
}

\Return $\mathcal{K}$\;
\end{algorithm}

\subsection{Evaluation Metrics}

\subsubsection{\textbf{VLM-based Evaluation Metrics}}
To comprehensively evaluate the quality of generated try-on videos, we introduce three VLM-based metrics: \textit{Appearance \& Background Consistency} (ABC), \textit{Garment Consistency} (GC), and \textit{Overall Visual Quality} (OVQ). Inspired by recent studies~\cite{lin2025dreamfit,peng2024dreambench++} showing strong alignment between large multimodal models and human judgments, we design a structured evaluation pipeline using Gemini 3.1 Pro as the evaluator.

\begin{figure*}[t]
    \centering
    \includegraphics[width=0.98\linewidth]{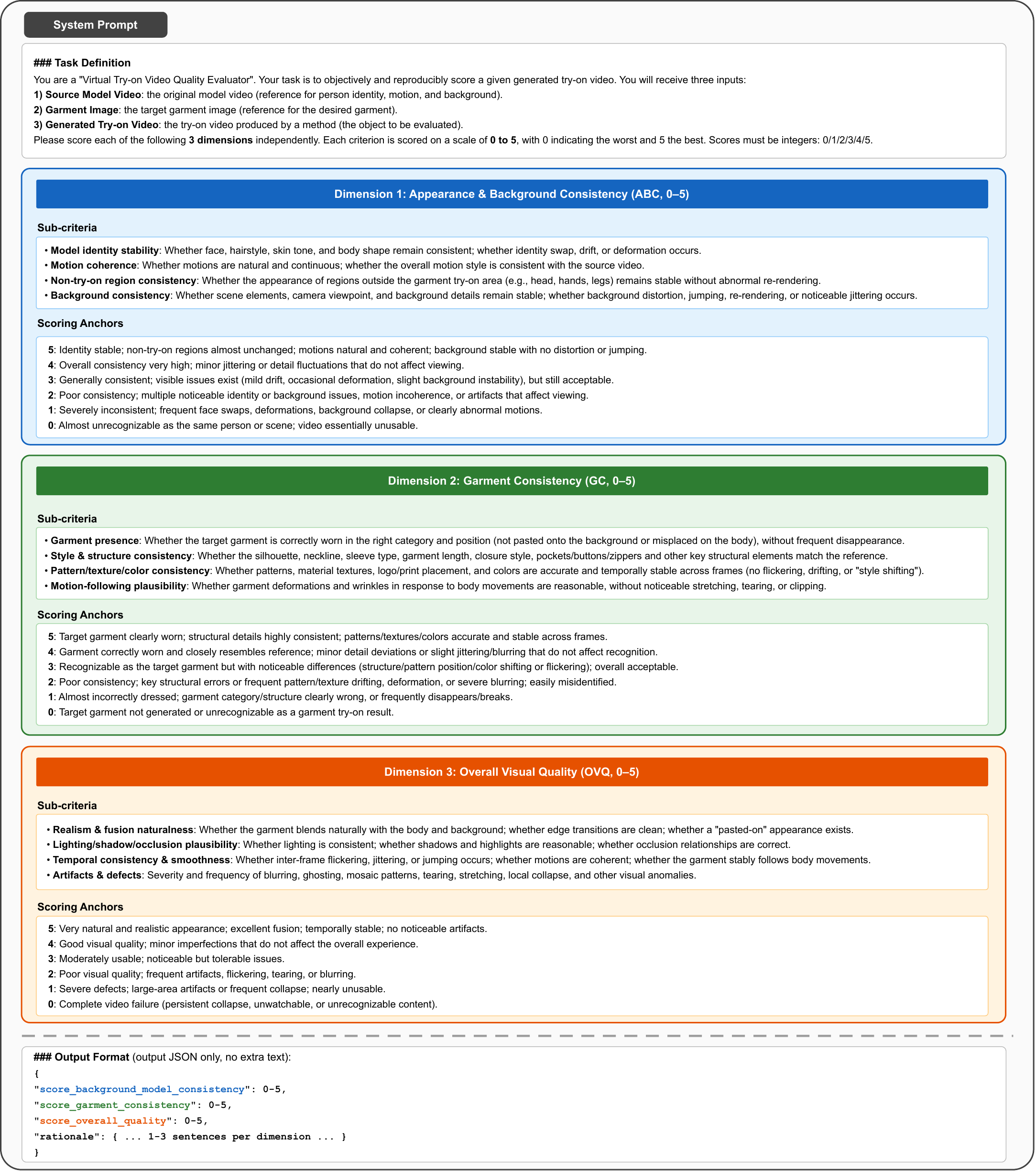}
    \caption{Evaluation prompt used for VLM-based video virtual try-on assessment. The evaluator is instructed to score each generated try-on video along three dimensions: Appearance \& Background Consistency (ABC), Garment Consistency (GC), and Overall Visual Quality (OVQ), each on an integer scale from 0 to 5 with detailed scoring anchors.}
    \label{fig:eval_prompt}
\end{figure*}

For each test sample, the evaluator receives three inputs: (1) the source model video, which serves as the reference for person identity, motion, and background; (2) the target garment image; and (3) the generated try-on video to be evaluated. The evaluator is instructed to assess the three dimensions independently, each on an integer scale from 0 (worst) to 5 (best). To ensure consistent and reproducible scoring, we provide detailed sub-criteria and scoring anchors for each dimension in the evaluation prompt, as shown in Fig.~\ref{fig:eval_prompt}. We set the decoding temperature to 0 and require the output to follow a structured JSON format containing the three integer scores together with brief rationales for each dimension.

We further validate the reliability of the VLM-based evaluation through a human study, which shows strong agreement between Gemini-based and human judgments; detailed results are provided in Appendix~\ref{sec:human_evaluation}.

\subsubsection{\textbf{Discussion on VFID Evaluation}}
\label{sec:supp_vfid}
Besides the FVD metric used in the main paper, prior VVT works have also widely reported the Video Fr\'{e}chet Inception Distance (VFID)~\cite{dong2019fw} for quantitative comparison. Although we do not use VFID as a primary metric in the main paper, we include it here for completeness and comparability with prior work.

During our study, we identified an implementation issue in a commonly used open-source VFID toolkit\footnote{\texttt{https://github.com/ZhengJun-AI/vfid-metrics}, \texttt{fid\_metrics/dataset.py}, lines 106--139.} adopted by several recent methods~\cite{chong2025catv2ton,li2025magictryon}. Specifically, regardless of the video length, the current implementation evaluates only the first 10 frames of each video and ignores all subsequent frames. As a result, the metric cannot fully assess quality over the entire video sequence, especially for videos with large motion variation or rapidly changing content. We provide a detailed analysis and a corrected implementation
on our project page. Based on the corrected implementation, we report VFID scores for all compared methods in Sec.~\ref{sec:additional_results}.

\subsection{Pseudo-Label Data Construction}
\label{sec:supp_pseudo}

As described in the main paper, we use two types of pseudo data in the multi-stage training strategy: image-level pseudo data for Stage~2 and video-level pseudo data for Stage~3. Their construction pipelines are detailed below.

\noindent\textbf{Image-level pseudo data.}
For each person image, we construct a corresponding pseudo image through the following steps. (1) We first use Qwen3-VL-32B~\cite{bai2025qwen3} to analyze the garment currently worn by the person and generate a garment editing instruction (e.g., ``change the striped t-shirt to a plain white hoodie''). (2) We then apply Qwen-Image-Edit-2511~\cite{wu2025qwen} to modify the garment region according to the instruction, producing a pseudo image. (3) Finally, we use Qwen3-VL-32B to verify whether the generated pseudo image is consistent with the editing instruction. Samples that fail verification are discarded. To accelerate inference, we adopt Qwen-Image-Edit-Lightning~\cite{qwen_image_edit_lightning_github}, a LoRA-based acceleration module for Qwen-Image-Edit. In total, we construct 30K pseudo images from person images covering diverse garment types, body poses, and real-world backgrounds.

\begin{table*}[t]
\centering
\caption{Per-sample construction cost of pseudo data. All timings are measured on a single NVIDIA A800 GPU.}
\label{tab:pseudo_cost}
\begin{tabular}{l|cc|cc|c}
\toprule
Data Type & Resolution & Frames & Image Editing / Try-on & Video Generation & Total \\
\midrule
Image & $832\!\times\!624$ & 1 & 4\,s & -- & 4\,s \\
Video & $832\!\times\!624$ & 65 & 12\,s & 262\,s & 274\,s \\
\bottomrule
\end{tabular}
\end{table*}

\noindent\textbf{Video-level pseudo data.}
Since existing open-source video editing models cannot produce sufficiently high-quality try-on videos, we use our Stage~1 model, BooM-VVT (Stage~1), to synthesize pseudo try-on videos. The construction pipeline is as follows. (1) Given a source video and a randomly selected target garment image, we apply GSKS to select keyframes from the source video. (2) We generate the corresponding keyframe try-on images using the multi-frame try-on model. (3) The source video, together with the keyframe try-on images and garment masks, is then fed into BooM-VVT (Stage~1) to generate a try-on video. (4) Finally, we use Qwen3-VL-32B to verify the generated video and ensure its quality. Samples with noticeable artifacts or garment inconsistencies are discarded. In total, we construct 2.5K video-level pseudo samples from unpaired human-centric videos.

\noindent\textbf{Cost analysis.}
Table~\ref{tab:pseudo_cost} compares the per-sample construction cost of the two data types. For image-level pseudo data, each sample requires only $\sim$4\,s for image editing. For video-level pseudo data, each sample requires $\sim$12\,s for keyframe try-on image synthesis and $\sim$262\,s for video generation, totaling $\sim$274\,s. Overall, constructing 30K image-level pseudo samples costs 33.3 GPU hours, while constructing 2.5K video-level pseudo samples costs 190.3 GPU hours. On a per-sample basis, image-level pseudo data is approximately $\mathbf{69\times}$ cheaper. This favorable cost ratio motivates our strategy of leveraging large-scale image-level pseudo data to expand data coverage at low cost, substantially reducing the reliance on expensive video-level pseudo data.

\section{More Results}
\label{sec:additional_results}
\subsection{Additional Quantitative Results}
\label{sec:vfid_results}

\begin{table}[t]
\centering
\caption{VFID evaluation using the corrected implementation.
$\text{VFID}_\text{I}$: I3D backbone; $\text{VFID}_\text{R}$: ResNeXt backbone.
$\downarrow$: lower is better. \textbf{Bold}: best; \underline{Underline}: second best.}
\label{tab:vfid_results}
\begin{tabular}{l|cc|cc}
\toprule
\multirow{2}{*}{Method} & \multicolumn{2}{c|}{ViViD-S} & \multicolumn{2}{c}{WildVVT} \\
 & $\text{VFID}_\text{I}\downarrow$ & $\text{VFID}_\text{R}\downarrow$
 & $\text{VFID}_\text{I}\downarrow$ & $\text{VFID}_\text{R}\downarrow$ \\
\midrule
ViViD~\cite{fang2024vivid}                    & 9.36  & \underline{0.20}  & 16.86 & \textbf{0.33} \\
CatV$^2$TON~\cite{chong2025catv2ton}          & 9.84  & 0.26              & 28.27 & 6.05 \\
MagicTryOn~\cite{li2025magictryon}            & 9.28  & 0.21              & 14.84 & 0.51 \\
BooM-VVT (Stage~1)                            & \underline{9.24}  & 0.24  & \underline{14.59} & 0.49 \\
BooM-VVT                                      & \textbf{9.16}  & \textbf{0.18}  & \textbf{13.96} & \underline{0.47} \\
\bottomrule
\end{tabular}
\end{table}

Using the corrected VFID implementation described in Sec.~\ref{sec:supp_vfid}, we report the VFID results on the ViViD and WildVVT datasets in Table~\ref{tab:vfid_results}. BooM-VVT achieves the best results in three of the four settings, and ranks second in the remaining one. In particular, on the more challenging WildVVT benchmark, BooM-VVT achieves the lowest $\mathrm{VFID}_{\mathrm{I}}$ by a clear margin, indicating stronger generalization in unconstrained real-world scenarios.

\subsubsection{Human Evaluation Validation.}
\label{sec:human_evaluation}
To further validate the reliability of our VLM-based evaluation,
we conduct a human evaluation on the WildVVT benchmark.
Specifically, we recruit 10 evaluators and use the same three
evaluation dimensions as in our VLM-based protocol:
Appearance \& Background Consistency (ABC), Garment Consistency
(GC), and Overall Visual Quality (OVQ). Each dimension is rated
on a 0--5 scale, and each generated video is independently
evaluated by at least two evaluators. The final score for each
method is obtained by averaging the corresponding human ratings.

As shown in Table~\ref{tab:human_eval}, BooM-VVT achieves the
highest human evaluation scores across all three dimensions. Furthermore, the human scores show strong agreement with our
Gemini-based evaluation, with an overall Pearson correlation of
approximately $r=0.9$.

\begin{table}[t]
\centering
\caption{Human evaluation on the WildVVT benchmark. Evaluators assess Appearance \& Background Consistency (ABC), Garment Consistency (GC), and Overall Visual Quality (OVQ)
on a 0--5 scale. Higher is better.}
\label{tab:human_eval}
\setlength{\tabcolsep}{5pt}
\begin{tabular}{lccc}
\toprule
Method& $ABC \uparrow$& $GC \uparrow$& $OVQ \uparrow$ \\
\midrule
ViViD & 3.48 & 2.57 & 2.62 \\
CatV2TON& 3.00 & 2.24 & 2.19 \\
MagicTryOn& 3.90 & 3.10 & 3.30 \\
BooM-VVT (Stage 1)& 4.48 & 3.95 & 4.14 \\
\textbf{BooM-VVT}& \textbf{4.71}& \textbf{4.24}& \textbf{4.43} \\
\bottomrule
\end{tabular}
\end{table}

\subsection{Additional Qualitative Results}

We provide additional qualitative comparisons on the WildVVT benchmark to further evaluate BooM-VVT under challenging real-world scenarios, including large motion, severe occlusion, and object interaction.

\begin{figure*}[ht]
    \centering
    \includegraphics[width=0.9\linewidth]{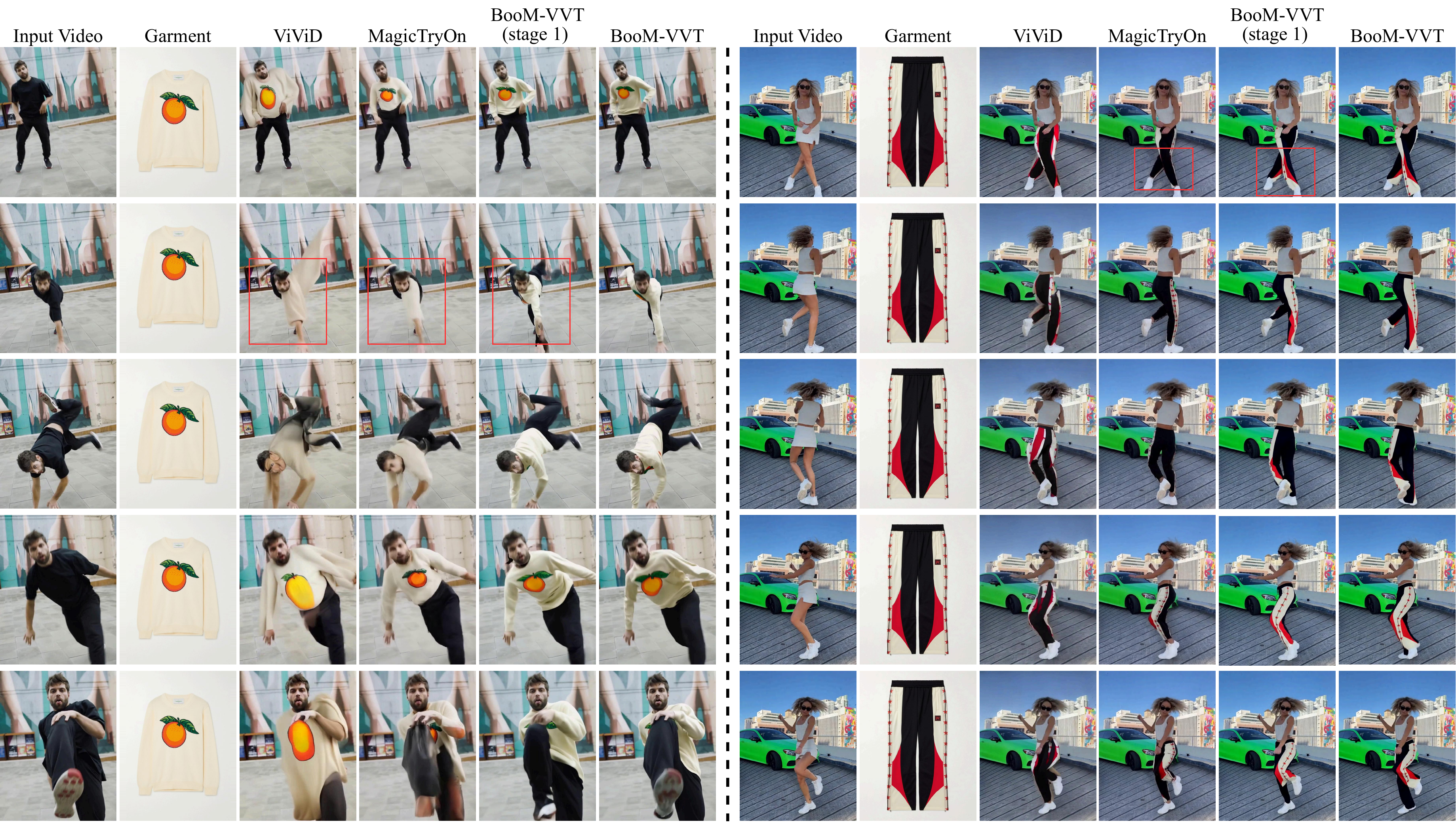}
    \caption{Qualitative results under large motion. Red boxes highlight artifacts produced by existing methods.}
    \label{fig:app_motion}
\end{figure*}

Fig.~\ref{fig:app_motion} presents qualitative results under large motion. Compared with existing methods, BooM-VVT produces more stable garment appearance and better temporal consistency. In the left example of Fig.~\ref{fig:app_motion}, the full mask-free model, BooM-VVT, also achieves better garment consistency than its mask-dependent variant, BooM-VVT (Stage~1), indicating that inaccurate masks caused by fast motion can degrade try-on region localization.

\begin{figure*}[t]
    \centering
    \includegraphics[width=0.95\linewidth]{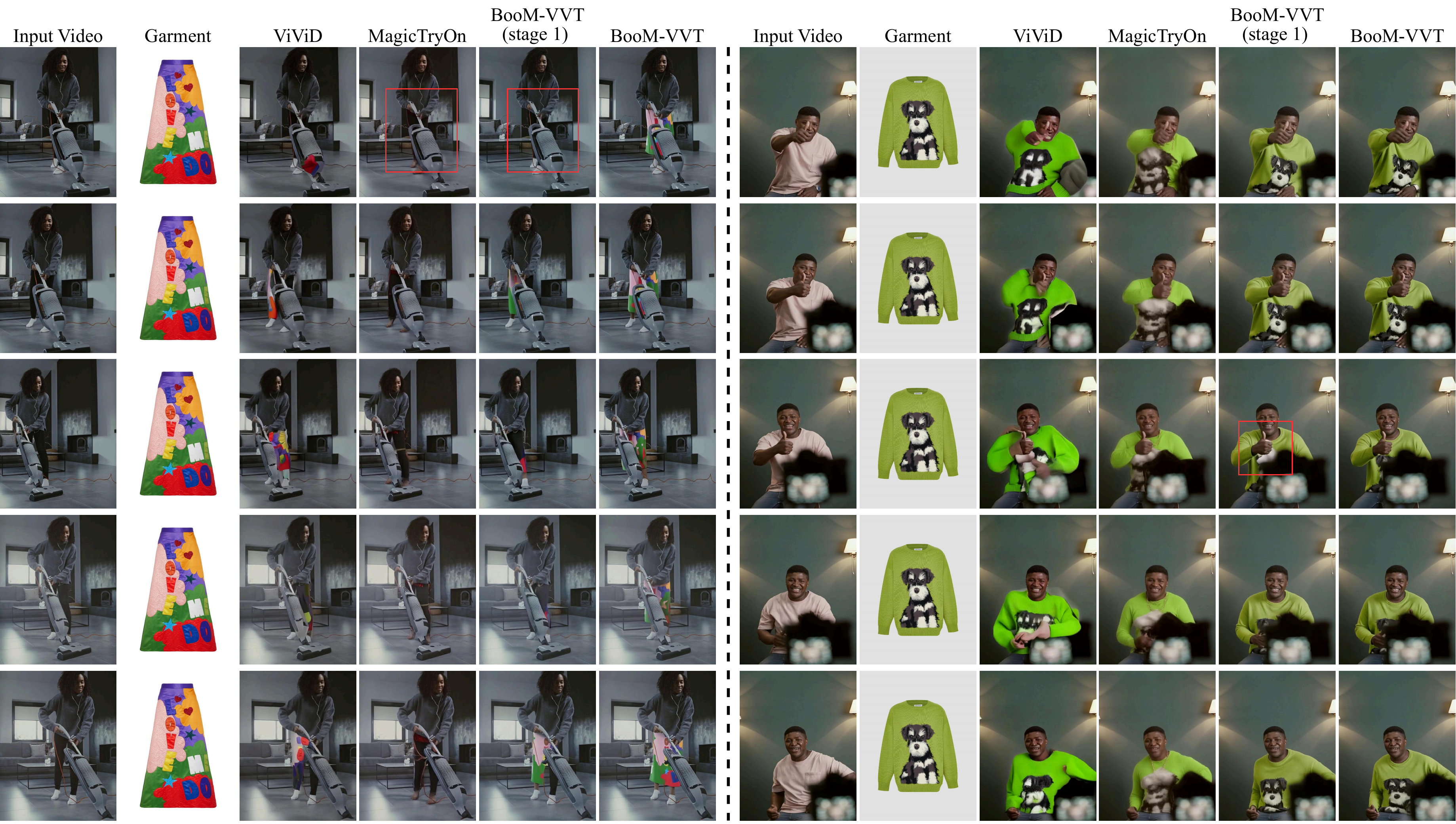}
    \caption{Qualitative results under severe occlusion. Red boxes highlight artifacts produced by existing methods.}
    \label{fig:app_occlusion}
\end{figure*}

Fig.~\ref{fig:app_occlusion} shows results under severe occlusion. In such cases, heavy self-occlusion makes garment transfer particularly challenging for existing methods, often leading to obvious artifacts or incorrect garment structures. By contrast, BooM-VVT maintains plausible garment appearance even when large body regions are occluded.

\begin{figure*}[t]
    \centering
    \includegraphics[width=0.95\linewidth]{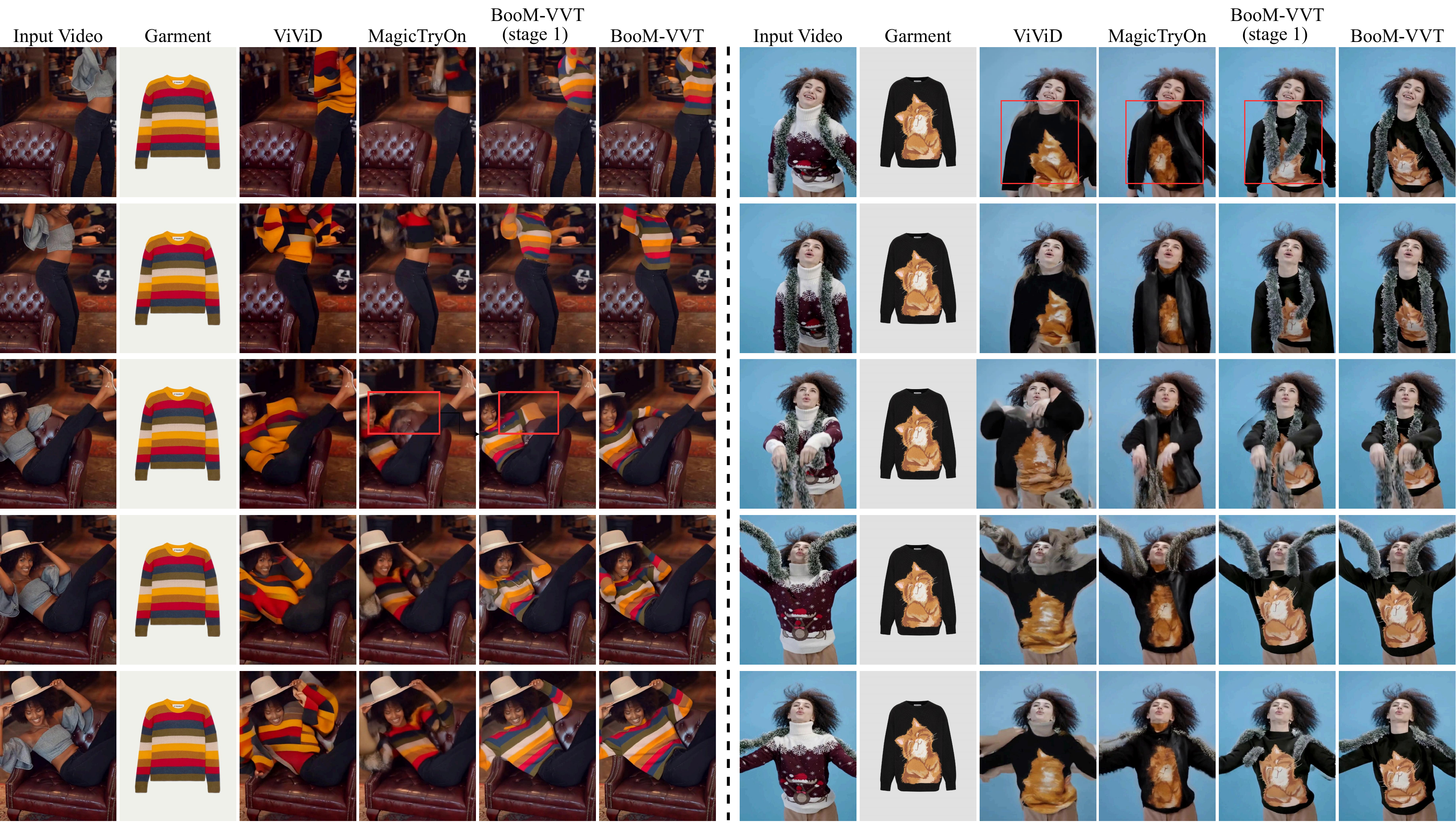}
    \caption{Qualitative results under object interaction. Red boxes highlight artifacts produced by existing methods.}
    \label{fig:app_interaction}
\end{figure*}

Fig.~\ref{fig:app_interaction} illustrates results in scenarios involving complex interactions between the person and surrounding objects. BooM-VVT better preserves the original interaction relationships, whereas existing methods often introduce noticeable artifacts in the interaction regions and may even distort or erase the interacting objects.

\subsection{Results on Diverse Try-on Tasks}

\begin{figure*}[t]
    \centering
    \includegraphics[width=0.85\linewidth]{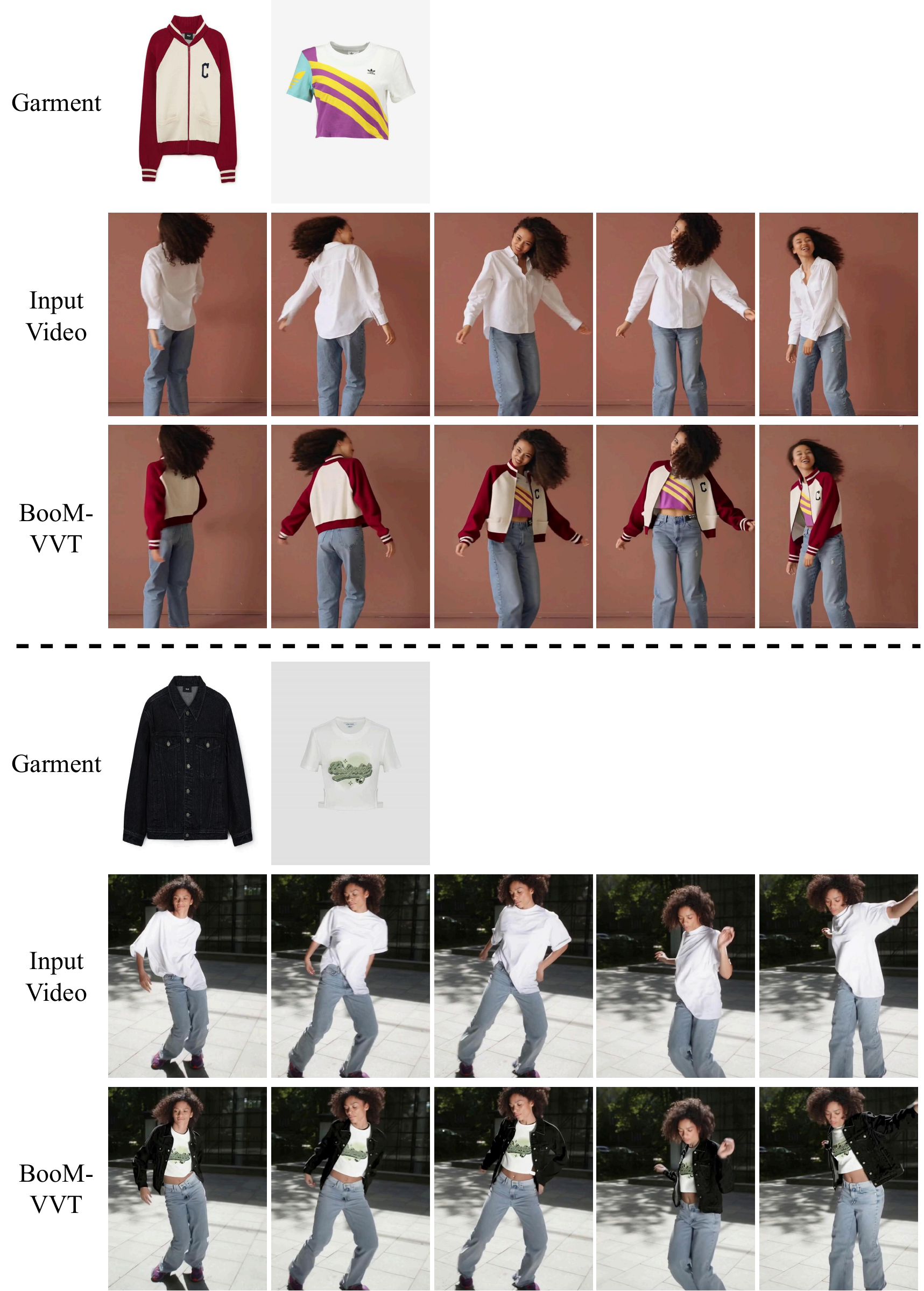}
    \caption{Results of BooM-VVT on the \textbf{layered try-on} task.}
    \label{fig:app_layer}
\end{figure*}

\begin{figure*}[t]
    \centering
    \includegraphics[width=0.85\linewidth]{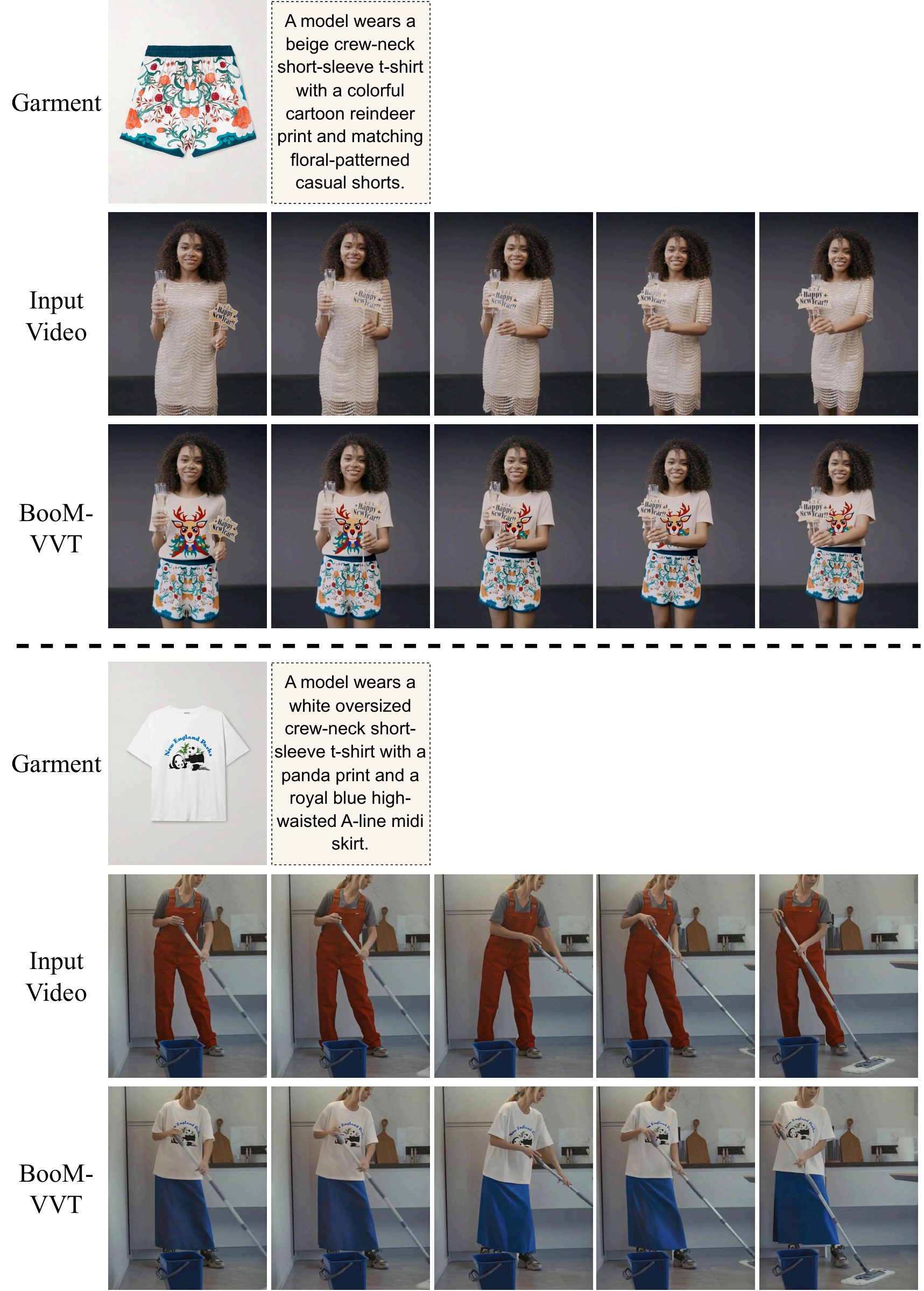}
    \caption{Results of BooM-VVT on the \textbf{cross-category try-on} task.}
    \label{fig:app_cross}
\end{figure*}

\begin{figure*}[t]
    \centering
    \includegraphics[width=0.69\linewidth]{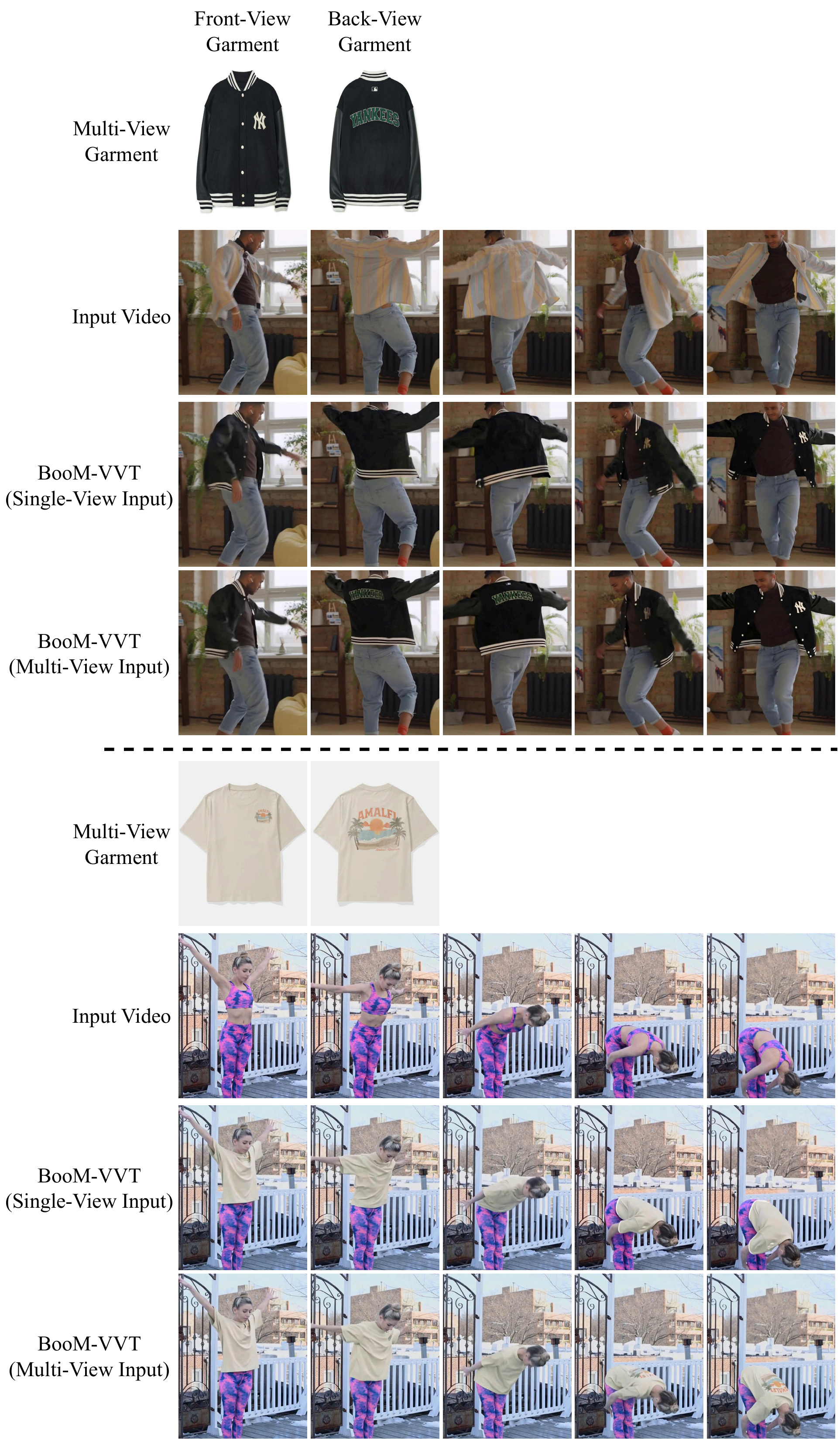}
    \caption{Results of BooM-VVT on the \textbf{multi-view try-on} task. The third and fourth rows show results with single-view (front only) and multi-view (front + back) garment inputs, respectively.}
    \label{fig:app_multiview}
\end{figure*}

Enabled by the proposed OmniView dataset, BooM-VVT supports diverse virtual try-on tasks, including layered try-on, cross-category try-on, and multi-view try-on. Fig.~\ref{fig:app_layer}, Fig.~\ref{fig:app_cross}, and Fig.~\ref{fig:app_multiview} show representative results for these three settings, demonstrating that BooM-VVT generalizes well across diverse try-on scenarios.

Specifically, for the \textbf{layered try-on} task (Fig.~\ref{fig:app_layer}), following DiOr~\cite{cui2021dressing}, we perform multiple keyframe try-on operations to sequentially dress garments layer by layer, from innerwear to outerwear. For the \textbf{cross-category try-on} task (Fig.~\ref{fig:app_cross}), we provide additional garment information in the text prompt to guide the model toward plausible cross-category try-on results. For the \textbf{multi-view try-on} task (Fig.~\ref{fig:app_multiview}), we concatenate garment images from different viewpoints and feed them into the multi-frame try-on model in a single forward pass. As shown in Fig.~\ref{fig:app_multiview}, BooM-VVT already produces reasonable results with only a front-view garment image. When multi-view garment images are additionally provided, the model generates more accurate results, especially for back-view frames.

\subsection{Additional Ablation Results}
\subsubsection{\textbf{Visualization of Garment-Sensitive Keyframe Sampling.}}
\label{sec:gsks_visualization}

\begin{figure}[t]
\centering
\includegraphics[width=0.9\columnwidth]{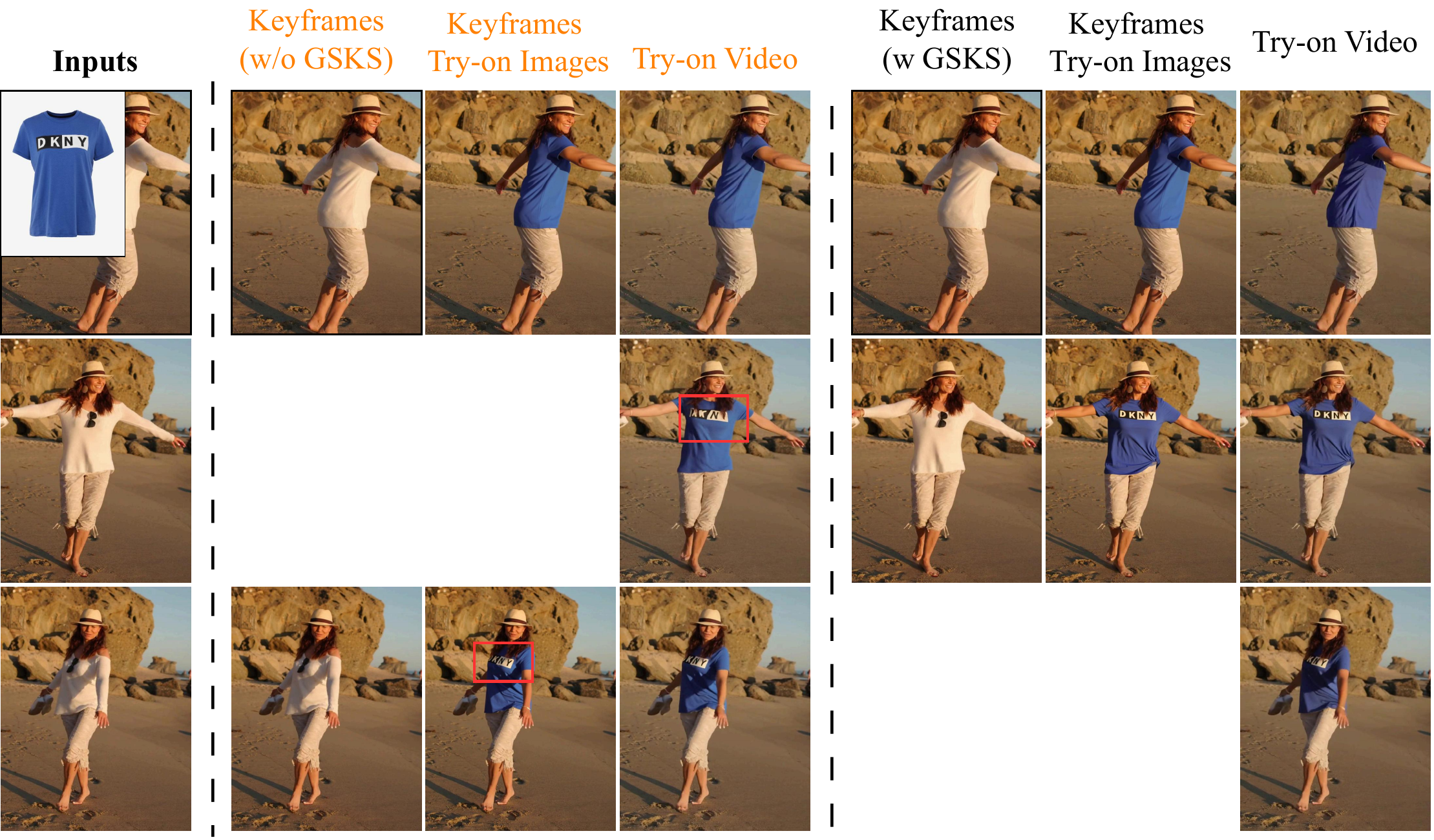}
\caption{Visual comparison of conventional keyframe sampling and GSKS.
GSKS focuses on garment-relevant body regions and selects more
informative keyframes, resulting in more faithful garment details
in both keyframe try-on images and final videos. Red boxes
highlight inconsistent garment details.}
\label{fig:gsks_vis}
\end{figure}

Fig.~\ref{fig:gsks_vis} provides a qualitative comparison between GSKS and conventional keyframe sampling based on global pose differences. Conventional sampling treats all body motions
equally, and therefore may select frames whose pose changes are mainly caused by garment-irrelevant body regions. Such keyframes provide limited appearance cues for the target garment and may further affect the quality of keyframe try-on and subsequent video generation.

In contrast, GSKS evaluates frame informativeness using garment-relevant body regions and additionally encourages complementary viewpoint coverage. This allows the selected keyframes to provide more effective spatial and viewpoint cues for garment appearance transfer. As shown in the figure, GSKS leads to more faithful reconstruction and preservation of fine-grained target-garment details, such as logos and patterns, in both the generated keyframe try-on images and the final try-on video. This qualitative comparison is consistent with the quantitative ablation results in Table~\ref{tab:ablation}, further validating the effectiveness of garment-sensitive keyframe selection.

\subsubsection{\textbf{Ablation on OmniView Dataset.}}
\begin{figure*}[t]
    \centering
    \includegraphics[width=1.0\linewidth]{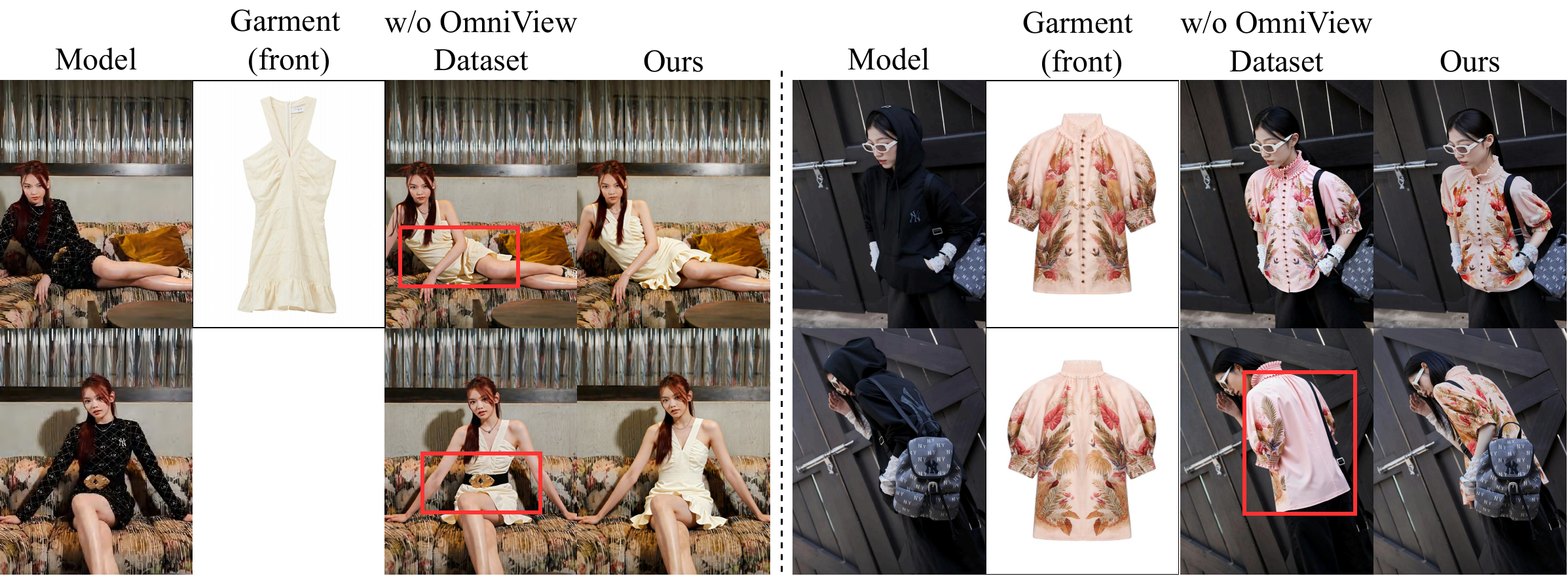}
    \caption{Qualitative ablation on the \textbf{OmniView dataset}. 
    \textbf{Left}: single-view garment input. The w/o OmniView 
    variant produces cross-view inconsistencies (e.g., the  belt highlighted by the red box). \textbf{Right}: 
    multi-view garment input (front \& back). The w/o OmniView 
    variant generates incorrect textures in back-view frames, 
    while the full model preserves accurate garment details. 
    Please zoom in for details.}
    \label{fig:abl_omniview}
\end{figure*}

To validate the effectiveness of the proposed OmniView dataset, we compare the full model with a variant whose multi-frame try-on model is not fine-tuned on OmniView (w/o OmniView). As shown in Fig.~\ref{fig:abl_omniview}, training with OmniView brings two key improvements.

First, OmniView enhances cross-view consistency in 
keyframe try-on results. As illustrated in the left example of Fig.~\ref{fig:abl_omniview}, the \emph{w/o OmniView} variant generates inconsistent garment details across viewpoints, such as a belt that appears in one view but disappears in another. In contrast, the full model produces 
consistent try-on results across viewpoints.

Second, OmniView significantly improves the model's capability 
for multi-view garment inputs. As shown in the right example of Fig.~\ref{fig:abl_omniview}, even when both front-view and back-view garment references are provided, the \emph{w/o OmniView} variant still produces incorrect garment textures when the person turns around. In contrast, the full model synthesizes sharper textures and more accurate garment structures that are consistent with the reference, highlighting the importance of high-quality and diverse multi-view try-on data.

\subsubsection{\textbf{Sensitivity to the weighting factor $\lambda$}}
We further study the effect of the weighting factor $\lambda$ in GSKS, which balances limb visibility and region visibility in the information-capacity score. Following the keyframe-driven setting, we evaluate BooM-VVT (Stage~1) on the WildVVT benchmark by reconstructing videos conditioned on the selected keyframes. The results are shown in Table~\ref{tab:lambda_ablation}.

Overall, the performance remains stable across a wide range of $\lambda$ values. In most cases, the model achieves very similar SSIM and LPIPS scores, indicating that GSKS is not overly sensitive to this hyperparameter. In particular, when $\lambda$ is set between 0.2 and 0.6, the results are consistently strong, with SSIM staying at around 0.871 and LPIPS at around 0.062. When $\lambda$ becomes too large, performance starts to degrade slightly, suggesting that overemphasizing region visibility may weaken the contribution of limb-based viewpoint cues. Based on this observation, we set $\lambda=0.2$ in all experiments.

\begin{table}[t]
\centering
\caption{Sensitivity analysis of the weighting factor $\lambda$ in GSKS on the WildVVT benchmark.}
\label{tab:lambda_ablation}
\begin{tabular}{c|cc}
\toprule
$\lambda$ & SSIM$\uparrow$ & LPIPS$\downarrow$ \\
\midrule
0.1 & 0.865 & 0.067 \\
0.2 & \textbf{0.871} & \textbf{0.062} \\
0.3 & \textbf{0.871} & \textbf{0.062} \\
0.4 & \textbf{0.871} & \textbf{0.062} \\
0.5 & 0.870 & \textbf{0.062} \\
0.6 & \textbf{0.871} & \textbf{0.062} \\
0.7 & 0.869 & 0.064 \\
0.8 & 0.867 & 0.069 \\
0.9 & 0.862 & 0.072 \\
\bottomrule
\end{tabular}
\end{table}

\end{document}